\documentclass{article}

\usepackage{iclr2027_conference,times}
\iclrfinalcopy

\usepackage[utf8]{inputenc}
\usepackage[T1]{fontenc}
\usepackage{hyperref}
\usepackage{url}
\usepackage{booktabs}
\usepackage{amsfonts}
\usepackage{amssymb}
\usepackage{amsmath}
\usepackage{nicefrac}
\usepackage{microtype}
\usepackage{xcolor}
\usepackage{graphicx}
\usepackage{subcaption}
\usepackage{tikz}
\usepackage{adjustbox}
\usepackage{multirow}
\usepackage{pifont}
\usepackage{wrapfig}

\usepackage{xspace}
\newcommand{\bench}{\textit{ForagerSim}\xspace}
\newcommand{\scenario}[1]{\texttt{#1}}

\definecolor{unvcolor}{RGB}{200,30,30}
\definecolor{editcolor}{RGB}{180,130,0}

\title{ForagerSim: Evaluating Embodied Map-Based Reasoning in Large-Scale Dynamic Worlds}

\author{%
\hspace*{-\tabcolsep}\begin{minipage}[t]{\textwidth}\centering
Tom\'{a}\v{s} Musil\textsuperscript{1} \quad
Martin Kot\'{i}k\textsuperscript{1} \quad
Michal Vavre\v{c}ka\textsuperscript{2} \\
Karla \v{S}t\v{e}p\'{a}nov\'{a}\textsuperscript{2} \quad
Sebastian Scherer\textsuperscript{3} \quad
Martin Saska\textsuperscript{1} \\[4pt]
{\normalfont
\textsuperscript{1}Dept. of Cybernetics, FEE, Czech Technical University in Prague \\
\textsuperscript{2}CIIRC, Czech Technical University in Prague \\
\textsuperscript{3}Robotics Institute, Carnegie Mellon University \\
\texttt{musilto8@fel.cvut.cz}}
\end{minipage}\hspace*{-\tabcolsep}%
}

\begin{document}
\maketitle
\lhead{Preprint, under review.}
\begin{abstract}
Animals build cognitive maps of vast environments, use them to perform map-based navigation, and might even be reusing the same circuitry used in spatial reasoning for creating and using maps of arbitrary task-spaces.
 In contrast, current embodied AI systems either rely on hand-engineered mapping pipelines, which break in the presence of unmodeled dynamics and  are restricted to metric space, or on learned policies that rarely form map-like representations at all.
Developing architectures that could learn to build and use map-like representations as flexibly as animals is, however, difficult because few 
  embodied AI benchmarks actually require map-based reasoning.
  To address this, we present \bench --- an interdisciplinary research platform and simulation environment focused on embodied tasks that require map-based reasoning, 
  with scenarios ranging from mobile-robot search and rescue to animal foraging in both complex mazes and vast, potentially infinite open areas.
  \bench is highly modular and allows modifying independent axes of difficulty related to perception (abstract 2D sensing to full RGB or LiDAR), scale (room-scale to multi-km scale), topology (open-fields to complex mazes), dynamics (smoke, debris, sensorimotor failures and others) and tasks (from simple 2D coverage to foraging-inspired well-alternation tasks).
  We compare reinforcement learning baselines with a simple engineered robotics stack and a human player on 8 example worlds which show how different difficulty axes affect existing methods.
  Our results show that, when sensing is simplified to 2D, learned policies can perform multi-object search and well-alternation tasks surprisingly well in simple topologies, but they fall short of human performance in larger or more complex-topology environments where map-based baselines perform well, suggesting that map-based reasoning is the bottleneck in these scenarios.
  Code and data are made open source \footnote{\href{https://github.com/MrTomzor/foragersim}{https://github.com/MrTomzor/foragersim}}.
\end{abstract}

\begin{figure}[htb]
\def\subfigwidth{0.24\linewidth}
\centering
\begin{minipage}{\subfigwidth}
  \begin{tikzpicture}
    \node[anchor=south west,inner sep=0] (a) at (0,0) {
      \adjustbox{trim={{0.05\width} {0.4\height} {0.6\width} {0.1\height}}, clip, width=1.0\linewidth}{%
        \includegraphics{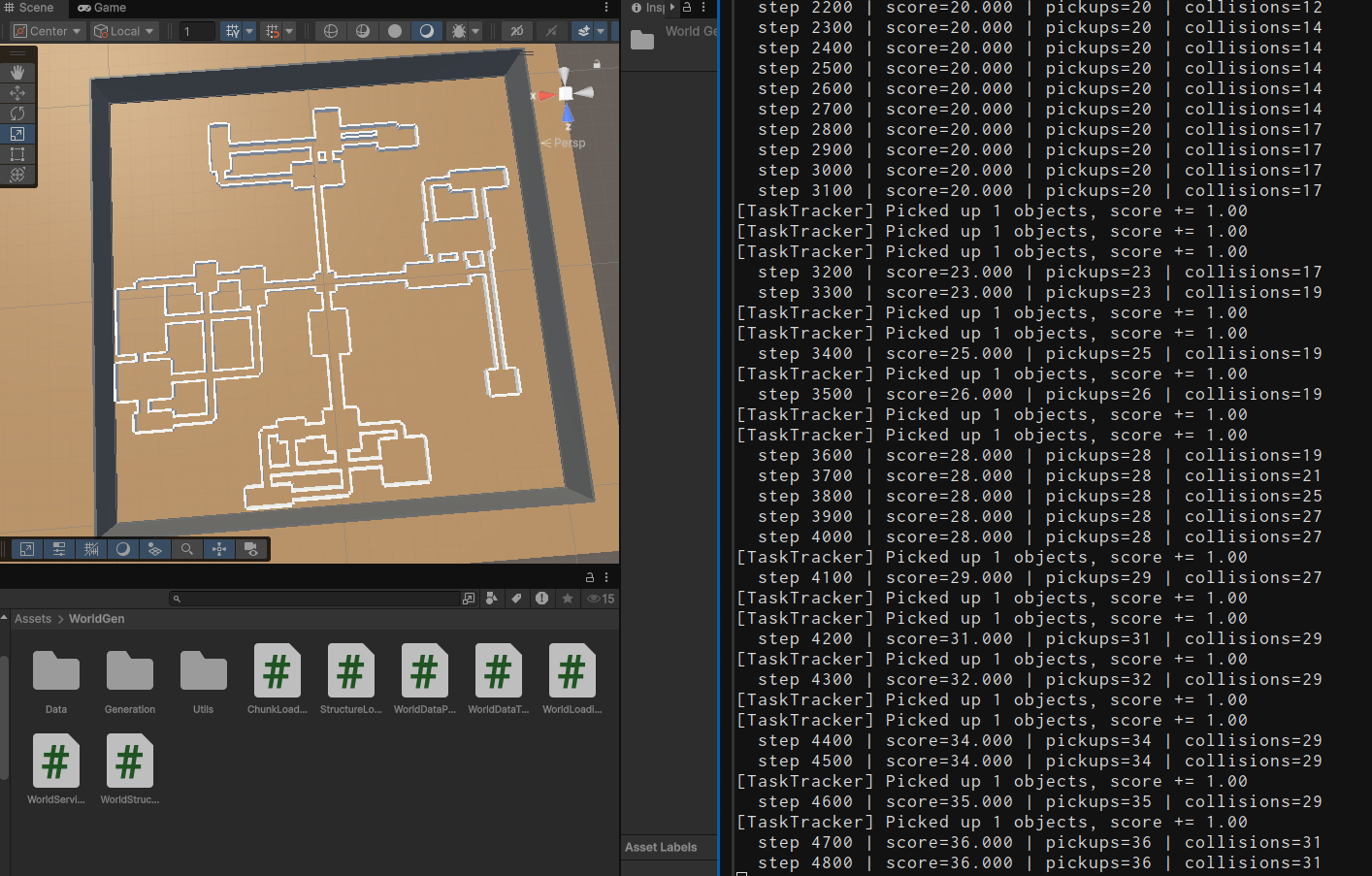}}};
  \end{tikzpicture}
\end{minipage}\hfill
\begin{minipage}{\subfigwidth}
  \begin{tikzpicture}
    \node[anchor=south west,inner sep=0] (a) at (0,0) {
      \adjustbox{trim={{0.0\width} {0.0\height} {0.0\width} {0.0\height}}, clip, width=1.0\linewidth}{%
        \includegraphics{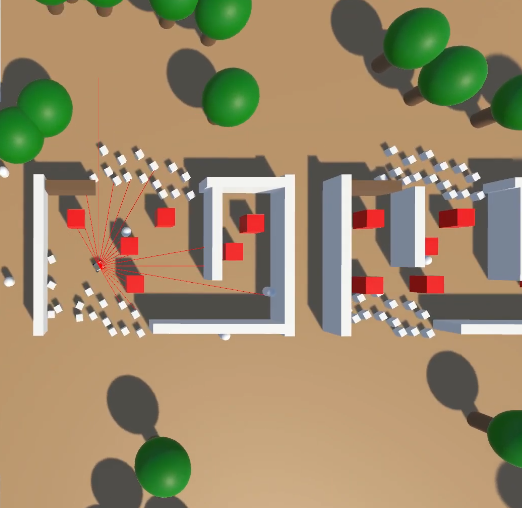}}};
  \end{tikzpicture}
\end{minipage}\hfill
\begin{minipage}{\subfigwidth}
  \begin{tikzpicture}
    \node[anchor=south west,inner sep=0] (a) at (0,0) {
      \adjustbox{trim={{0.0\width} {0.0\height} {0.0\width} {0.0\height}}, clip, width=1.0\linewidth}{%
        \includegraphics{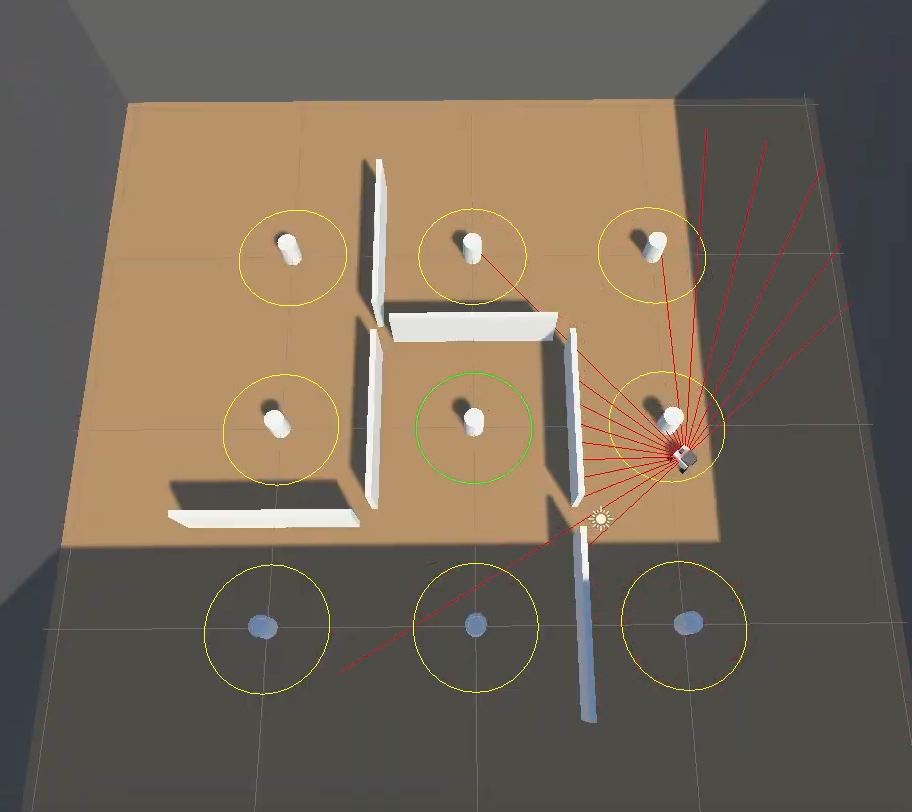}}};
  \end{tikzpicture}
\end{minipage}\hfill
\begin{minipage}{\subfigwidth}
  \begin{tikzpicture}
    \node[anchor=south west,inner sep=0] (a) at (0,0) {
      \adjustbox{trim={{0.2\width} {0.0\height} {0.2\width} {0.0\height}}, clip, width=1.0\linewidth}{%
        \includegraphics{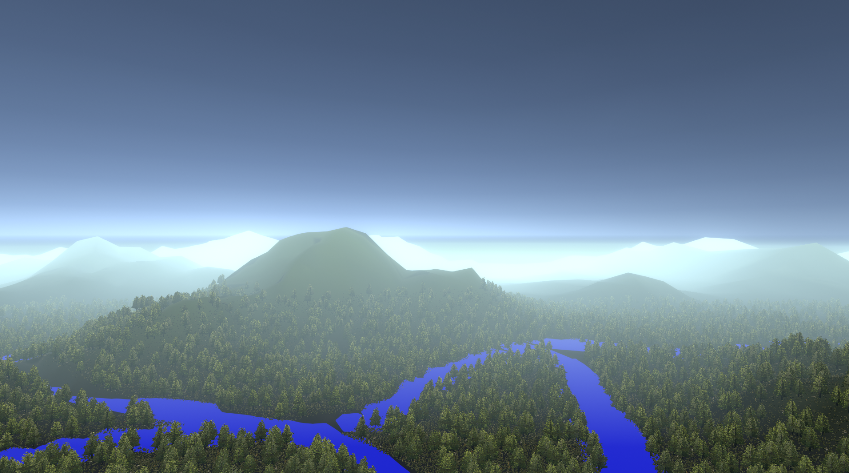}}};
  \end{tikzpicture}
\end{minipage}
  \caption{Example scenarios in \bench - from left to right: The \scenario{4arm-submazes} maze; multi-object search in a dynamic-environment consisting of houses with movable rubble and chaotically walking entities; \scenario{ratmaze-foraging} experiment with rewards alternating between random wells and a home well; massive-scale realistic bat-level foraging scenario}
\label{fig:hero}
\end{figure}

\section{Introduction}
\label{sec:intro}
Searching for sparse objects (e.g. food) in vast, dynamic environments (deformable and changing), and then returning to known locations (e.g. a nest, a storage, or favorite feeding locations) are critical capabilities for the survival of nearly all animal species.
  When performing this task in large environments, animals and humans do not behave purely reactively, but they build highly scalable spatial representations (cognitive maps) \citep{tolmanCognitiveMapsRats1948, Epstein2017TheCM, gevasagivSpatialCognitionBats2015} which enable efficient planning, navigation, and reacting to environment changes.
Bats and other animals are capable of navigating at scales as high as tens of kilometers \citep{kidnapped_bats, wolves_navigation}.
Dogs trained for search and rescue missions learn to navigate rubble, enter damaged structures, and search systematically over large areas through a curriculum of reinforcement, without being told how to map or plan~\citep{gerritsenK9SearchRescue2014}, and without human-level language understanding.
Some research even suggests that the cognitive map construction could play a significant role in general task learning in the form of learning reusable \textit{schemata} that agents can then use for faster learning, efficient decision making, or imagining future trajectories in both physical and abstract spaces \citep{aronovMappingNonspatialDimension2017, baraducSchemaCellsMacaque2019, Epstein2017TheCM}.
Importantly, replicating the learnable map-based reasoning observed in animals and humans could enable agents to learn to perform tasks that are currently unsolvable by neither hand-crafted map-based reasoning systems nor by existing end-to-end learning approaches by combining the strengths of both.
The main motivating example, which we also use as inspiration in designing some of the dynamics and topologies in \bench, are search-and-rescue missions in environments that are both large, and highly dynamic (e.g. autonomously finding survivors in a suburb during a wildfire, flood or earthquake), which have not yet been reliably solved by any existing embodied AI approach.



Engineered mobile robotics systems, often called autonomy stacks \citep{darpa_cerberus_wins, csiro_darpa, beneath}, serve as examples showing that performing localization, mapping and planning at multiple levels of abstraction (e.g. using a global+local planner, and combining both metric and topological maps) can allow robots to search and navigate unknown environments at scales of even up to several kilometers.
However, they are also built on assumptions that break in highly dynamic environments, since every new type of dynamics has to be engineered into the pipeline by hand (e.g. how the mapping and plannning has to be changed, compared to a static environment, in the case of smoke, humans or e.g. actuator or odometry failures).
Even though deep learning is sometimes used in parts of these systems (e.g. for perception or low-level control), the mapping and planning modules, their interactions and their hierarchy levels are hand-designed and not learnable end-to-end.

While deep learning approaches can learn to perform arbitrary tasks in worlds with arbitrary dynamics, for which handcrafting autonomy stacks would be completely infeasible, map-like representations and online planning capabilities are severely limited in existing learning architectures. 
Some recent works have investigated the map-building capabilities of commonly used learned memory architectures, and found that map-like representations can emerge, but only in highly specific conditions, \citep{wijmansEmergenceMapsMemories2023, banino2018vector} or when a structural prior on spatial transformations is embedded into the architecture \citep{yangSpatiallyenhancedRecurrentMemory2025}.
While recent research has shown that the popularly used large language models can sometimes provide reasonable estimates on ``where to go to search for a given object'',  studies have shown that they are very brittle when tested on long-horizon embodied tasks that require spatial reasoning~\citep{balrog}.
Some approaches have combined classical autonomy stacks with taking LLM outputs as high-level guidance, which has had success in static environments~\citep{raven}, but this approach still suffers from the core problems of engineered autonomy stacks (inability to learn and brittleness to noise and hard-to-model situations).


Finding learning architectures that could learn to build and utilize maps at the scales and flexibility seen in animals could benefit the robotics community (by removing the need to re-engineer mapping/planning pipelines and heuristics for different morphologies and environments), the reinforcement learning community (by helping with long-term memory and behavior), and the cognitive science community (by providing testable models of how map-like representations may arise in animals).
However, each community evaluates these capabilities on its own tasks, under its own assumptions: robotics on realistic sensing in largely static worlds, reinforcement learning on memory tasks in small abstract environments, and cognitive science on animal experiments that are rarely reproduced in simulation and often only in small box-arenas.
To provide a platform for these communities to collaboratively investigate the problem of learnable map-based reasoning, our paper brings the following contributions: 

\begin{itemize}
  \item \textbf{Environment} 
We introduce ForagerSim, an open-source environment for embodied search and navigation in which perception, scale, topology and dynamics can be varied independently, allowing the construction of tasks that truly require map-like memory and online planning. 
To keep it accessible, we design the worlds such that the same scenarios can be addressed in both a fast, low-dimensionality abstract 2D depth+semantics observation mode and a more detailed, 3D mode with RGB/LiDAR sensing.
    We replicate scenarios from mobile robot navigation, memory-learning benchmarks and animal experiments, design new environmental topologies and dynamics which challenge existing methods and present an example curriculum learning experiment which smoothly increases topological complexity.

  \item \textbf{Baseline RL Scores} We benchmark a model-free and a model-based agent on 8 world configurations along 3 axes that isolate topology, dynamics, and task complexity, and find that while RL baselines show some learning on simple worlds, they are easily broken by more complex world topologies and by the more map-focused well-alternation task. 
  \item \textbf{Comparison with a Classic Mobile-Robotics Exploration Stack} For the multi-object search settings, we compare learned approaches with a simple engineered autonomy stack implemented in ROS2 and note its high performance on complex topologies and failure cases in worlds with dynamics or non-uniform distributions of reward objects. 
  \item \textbf{Human Reference} We record the performance of a human player and provide a robotics-inspired UI that makes the comparison fair when simulating in the fast, 2D raycasting mode. This offers an estimate of the gap between current methods and human performance.
\end{itemize}

\section{Related Work}
\label{sec:related}
\label{sec:related_work}
In this section, we summarize relevant existing benchmarks and simulation environments and discuss their limitations which have motivated the development of \bench.
Table~\ref{tab:sim-comparison} summarizes how \bench differs along the axes most relevant to evaluating map-based reasoning.

\paragraph{Robotics Simulation Environments}
The most popular embodied-AI simulation environments, such as Habitat~\citep{habitat, habitat3} and ProcTHOR~\citep{procthor}, allow simulating embodied agents in photorealistic scanned or generated household interiors, but their worlds are limited to the scale of a single building and mostly static.
ExploreBench~\citep{explore_bench} is focused specifically on benchmarking exploration approaches, but its worlds are static and also limited to indoor layouts. 
Outdoor and large-scale simulators, such as AirSim~\citep{airsim}, UnrealZoo~\citep{unrealzoo} or MetaUrban~\citep{metaurban}, offer simulation of various robot morphologies in large, photorealistic worlds, including urban scenes with pedestrians.
However, their high rendering cost makes RL training expensive (e.g. UnrealZoo reaches 83 FPS on an RTX~4090).

\paragraph{Long-Horizon Behavior RL Benchmarks}
MemoryMaze \cite{memorymaze} simulates ordered multi-object search tasks in mazes and focuses on RL memory evaluation.
We replicate the same environment topology and scale in \scenario{memorymaze15x15} and show that for the task of unordered object-search, even the largest maze can easily be solved with memoryless policies.
MemoryGym \cite{Pleines2023MemoryGT} is a newer benchmark focusing on memory on multiple videogame-like scenarios, which can be endless, but are constrained to abstract grid-world scenarios.
POPGym~\citep{popgym} contains some grid-world maze-navigation scenarios, but the authors state that wall-following memoryless policies were able to solve these, suggesting that navigation on its own might not be a strong memory benchmark.
Videogame RL wrappers like Crafter~\citep{crafter}, Craftax~\citep{craftax}, NetHack~\citep{nethack}, and MiniHack~\citep{minihack} offer rich open-ended videogame worlds that test long-horizon behavior, but are constrained to highly abstract grid-world scenarios.
Minecraft-based suites like MineRL and MineDojo~\citep{minerl, minedojo} do provide infinite 3D worlds, but they inherit the slowness of Minecraft and the observations and actions are still highly abstract compared to mobile robot embodiments.

\paragraph{Animal-Inspired Environments}
Some simulation environments also take inspiration from specific animal capabilities and pose those as learning challenges.
Avalon~\citep{avalon}, focused on RL generalization, provides procedural 3D videogame-style island worlds with 20 embodied tasks sharing dynamics and action space. 
Its scenes range up to 600m in scale, but they contain simple topologies where the reward area can often be seen from the start, and the observation and action dimensionality are relatively high, due to being designed as VR-compatible.
Animal-AI~\citep{animal_ai_bench} recreates 900 tasks from cognitive science, such as object permanence, radial-arm-maze tasks, but all within a 40x40 units wide arena, which is not suitable for large-scale navigation.

Compared to the existing environments, \bench is the only one that combines massive scalability (through Minecraft-inspired world generation and streaming), modularity (specifically in being able to independently change axis of difficulty of topology, scale, observations, dynamics and switching between 2D and 3D variants of the same scenario), and cross-disciplinary focus (through offering both a ROS2 and Gym interface and by presenting example environments replicated from across mobile robotics, memory learning benchmarks and animal studies) which are necessary for investigating the problem of learning how to map, plan and navigate at the level of freely foraging animals.

\newcommand{\yes}{\checkmark}
\newcommand{\no}{\ding{55}}
\newcommand{\partmacro}{$\sim$}
\newif\iftablecites
\tablecitesfalse 
\newcommand{\tabcite}[1]{\iftablecites~\citep{#1}\else\nocite{#1}\fi}
\begin{table*}[t]
\centering
\small
\setlength{\tabcolsep}{4pt}
\renewcommand{\arraystretch}{1.15}
\makebox[\textwidth][c]{%
\begin{tabular}{l l c c c l}
\toprule
\textbf{Simulator} & \textbf{Worlds} & \textbf{Dynamic} & \textbf{Infinite} & \textbf{Modular} & \textbf{Interface} \\
 & & \textbf{Environments} & \textbf{Scaling} & \textbf{WorldGen} & \\
\midrule
Explore-Bench\tabcite{explore_bench}          & 2D+3D cont. & \no  & \no  & \partmacro & ROS1 \\
Habitat / EAI suites\tabcite{habitat, habitat3} & 3D cont. & \partmacro & \no  & \yes       & Gym+ROS1 \\
NetHack / MiniHack\tabcite{nethack, minihack} & 2D grid   & \yes & \yes & \yes       & Gym \\
Craftax / Crafter\tabcite{craftax, crafter}   & 2D grid   & \yes & \no  & \partmacro & Gym \\
MineRL / MineDojo\tabcite{minerl, minedojo}   & 3D cont.  & \yes & \yes & \partmacro       & Gym \\
Memory Maze\tabcite{memorymaze}               & 3D cont.  & \no  & \no  & \no        & Gym \\
Animal-AI\tabcite{animal_ai_bench}            & 3D cont.  & \yes & \no  & \yes       & Gym \\
Avalon\tabcite{avalon}                        & 3D cont.  & \yes & \no  & \yes       & Gym \\
AirSim\tabcite{airsim}                        & 3D cont.  & \partmacro & \no & \no   & Gym+ROS1 \\
UnrealZoo\tabcite{unrealzoo}                  & 3D cont.  & \yes & \no  & \no        & Gym \\
MetaUrban\tabcite{metaurban}                  & 3D cont.  & \yes & \no  & \yes       & Gym+ROS2 \\
\hline
\textbf{ForagerSim (ours)} & 2D+3D cont. & \yes & \yes & \yes & Gym+ROS2 \\
\bottomrule
\end{tabular}%
}
\caption{Comparison of related robotics simulators, RL environments and benchmarks involving long-horizon embodied tasks.
\yes~= implemented, \protect\partmacro~= partial or limited, \no~= absent.
\textbf{Worlds}: dimensionality of the simulated worlds and whether it is a discrete gridworld (grid) or continuous (cont.); \bench flattens its 3D worlds into a fast 2D mode with abstracted sensing.
\textbf{Infinite Scaling}: the simulation is designed so that step cost does not grow with world size, e.g.\ by generating and simulating only the region around the agents.
  }
\label{tab:sim-comparison}
\end{table*}

\section{The \bench Environment}
\label{sec:environment}
\bench is a domain of randomized 2D and 3D environments of various topologies, scales and dynamics designed for evaluating map-based reasoning in scenarios inspired by mobile robotics applications and animal studies.
The core simulation and procedural generation logic of \bench is implemented from scratch inside the Unity engine, which communicates with external agents via either an AI Gym~\citep{gymnasium} interface or a ROS\,2 interface, with messages routed through a lightweight usage-agnostic Python wrapper of the simulator shared by both interfaces.
In this section, we present some details about \bench and the design decisions that led to them.

\subsection{Modular Scenario Generation and Large-World Simulation}
\label{sec:world-defs}
In \bench, any given simulation scenario is fully defined by three JSON configuration files specifying the \textit{world} (procedural generation rules, presence of dynamics, object placements etc.), the \textit{agent(s)} (number of agents, sensors, actuators, and their noise models), and the \textit{task} (rewards and termination conditions).
Both the ROS2 and Gym interfaces use the same config files, and the simulation wrapper contains an independent tracker evaluating rewards and termination, which allows evaluating a Gym-trained RL policy and a ROS\,2 autonomy stack under identical environments and the same scoring.
These config files allow for easily creating new variations of existing scenarios, including the ability to define an adaptive difficulty curriculum that smoothly interpolates between values of any world-parameter (e.g. the arena size, number of rooms etc.), as visualized in \autoref{fig:allworlds_adaptive}.
With only a few lines of config-file changes, a user can easily create new scenarios and independently vary the following axes of difficulty:
\begin{itemize}
  \item \textbf{Environment topology and scale}: can vary from open-spaces and suburb-like configurations to mazes (including the MemoryMaze topology from \cite{memorymaze} and the rat-maze from \cite{widloskiFlexibleReroutingHippocampal2022}), with potentially infinite scales due to procedural world streaming only close to the agents' sensing radius.
  \item \textbf{Environment dynamics}: inspired by search-and-rescue missions, we implement movable rubble, openable or lockable doors, chaotically moving walkers and moving smoke which degrades depth perception.
  \item \textbf{Task complexity}: we allow independently defining distributions of static rewards (e.g. in specific structures or uniformly distributed), and also switching the complexity of reward obtaining from simple static pickupable objects to 'wells' that look identically from afar and the agent must get close to to receive rewards, as is done in some cognitive science expeirments \cite{widloskiFlexibleReroutingHippocampal2022}), 
  \item \textbf{Agents' sensing and actuation configurations}: While the worlds are fundamentally 3D, the environment is designed so that sensing can run in both a fast abstracted 2D depth and semantics sensing, which can run at more than 3000 steps per second on a standard GPU-less laptop, or a 3D mode with a 3D LiDAR or RGB/RGB-D cameras. Noisy IMU or perfect pose estimation can also be used and the simulation supports multi-agent definitions. 
\end{itemize}

Inspired by Minecraft's ability to generate and simulate infinite worlds, \bench only streams the world in an area near the agent, which enables simulating missions that span tens of kilometers in scale.
All procedural generation is deterministic given a seed, with each 'axis' (topology, dynamics, task) deriving their own seeds, ensuring reproducibility and modularity.
Implementation details of the world generation can be found in \autoref{sec:sim_details}.

\subsection{Reward Signal and Tasks}
To enable evaluating multi-task learning in future work, we use a unified reward function in all our presented experiment and only change the nature of the task in each scenario through the \textit{world} config.
In any world, the agent gets a +50 reward for touching a reward object (which then disappears),
a small negative reward -0.1 for colliding with other objects, and also a small reward of 0.01 for every square meter that it saw with its sensors, which helps in bootstraping learning, especially in cases where the reward objects or wells are extremely sparse.
\bench implements this spatial coverage reward by tracking the explored area covered by an agent as a 2D occupancy grid spanning the world, updated using the ground-truth pose and raycast data (even if the agent's own observations are noisy) as in traditional occupancy mapping approaches \cite{paper_frontier_grandpa}). 
We evaluate two principial task variations:

\paragraph{Multi-Object Search.} In these world configurations, the reward objects (red cubes, with different semantic classes than obstacles) are spawned in the world and only need to be found and touched to be obtained.
The spawn locations differ between worlds, but we keep a fixed maximum number of spawned objects per world config.
Since our results show that this task can be sometimes be solved with reactive approaches in simpler topologies, we also designed the following task:

\paragraph{Foraging (well alternation).} These world configrations have no freely placed reward objects, but rather they contain a variable number of \textit{wells}, which are static objects with different semantic classes than obstacles.
To obtain rewards from a well, the well must first be in a primed state (invisible to the agent from distance) and the agent must stand for a specified number of timesteps near the well, after which a number of reward objects is spawned next to the well and the agent can pick them up, just like in multi-object search.
In our experiments, we recreate the well-alternation pattern from \cite{widloskiFlexibleReroutingHippocampal2022}, where the agent starts next to a "home" well, and the well priming alternates between one randomly selected well, followed by the home well, ad infinitum.
The reasoning behind this pattern is that it simulates the scenario of an animal repeatedly going away from its home to search for food, and then returning to one known location (e.g. nest), which should benefit from building and keeping a map of the environment and is harder to solve with random-walk approaches than the basic multi-object search.
As in \cite{widloskiFlexibleReroutingHippocampal2022}, we also provide a directional signal for guiding the agent to the random wells, but the signal is turned off for the home well, forcing the agent to utilize map-based reasoning to return home efficiently. 

\subsection{Observations, Actions, and Embodiment}

\begin{wrapfigure}{r}{0.5\textwidth}
\def\subfigwidth{0.49\linewidth}
\centering
\begin{minipage}{\subfigwidth}
  \begin{tikzpicture}
    \node[anchor=south west,inner sep=0] (a) at (0,0) {
      \adjustbox{trim={{0.05\width} {0.03\height} {0.05\width} {0.0\height}}, clip, width=1.0\linewidth}{%
        \includegraphics{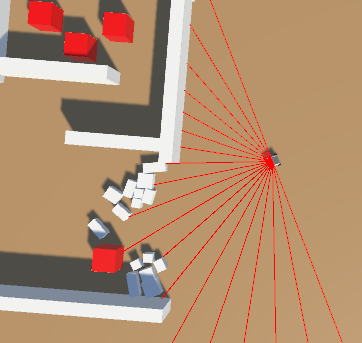}}};
  \end{tikzpicture}
\end{minipage}
\begin{minipage}{\subfigwidth}
  \begin{tikzpicture}
    \node[anchor=south west,inner sep=0] (a) at (0,0) {
      \adjustbox{trim={{0.05\width} {0.0\height} {0.05\width} {0.0\height}}, clip, width=1.0\linewidth}{%
        \includegraphics{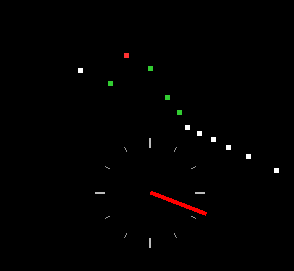}}};
  \end{tikzpicture}
\end{minipage}
  \caption{Illustration of the agent's observations. Left: top-down view of the environment and the raycasts. Right: human-UI visualization of the depth, semantics and compass data. }
\label{fig:obs_modes}
\end{wrapfigure}

\paragraph{Observation space.}
\bench supports a range of sensors common to mobile robotics: RGB and RGBD cameras, 2D lidar, 3D lidar, GPS, compass, and general odometry. To isolate long-horizon embodied behavior from perception while allowing fast simulation, we design and use an abstraction on sensing, which we dub the \emph{2D Semantic Depth Sensor} (2DSDS), visualized in Figure~\ref{fig:obs_modes}.
The observation is an $N \times (D+1)$ tensor, where $N$ is the number of rays cast in the horizontal plane (19 in our experiments) and each ray returns a distance normalized by the sensor's max range together with a $D$-dimensional one-hot class embedding. Language labels for classes are also available but not used in our experiments.
This abstraction is groundable in current mobile robot systems: depth sensors are standard in modern autonomy stacks, and semantic segmentation can be assumed to work robustly at moderate distances with modern foundation models~\citep{segment_anything}.
The 2DSDS sensing abstraciton allows the simulation to run at over 3000 steps per second on a standard CPU-only laptop, enabling large-scale RL experiments in large-scale dynamics worlds with millions of environment steps at negligible computational cost.
\label{sec:obs_act}
In the well-alternation task, we also give the agent cues that indicate the direction to the next random well (but never the home well).
We also allow simulating sensory failures and faults which are common in field robotics but often neglected in embodied AI research, such as odometry drift, lens occlusion and noise generated by smoke particles (used e.g. in the \scenario{5houses-sar} world).


\paragraph{Action space and embodiment}
We use an idealized velocity controller discretized into 3 linear velocity and 5 angular velocity bins. 
We also offer acceleration-based control and simulating noise in the controls, or even complete failures that could be encountered in real-world robotic missions such as the inability to turn one way or the inability to move backwards.
The agent is a spherical 1-unit wide object moving in the world, able to push dynamic objects with its movement.

\section{Experiments}
\label{sec:benchmark}
In this section, we present evaluations of multiple baseline approaches on the 8 world configurations described in \autoref{sec:scenario_desc}, and curriculum learning experiments on the hard-topology worlds in \autoref{sec:adaptive_experiment}.
We have evaluated the following RL baselines, a handcrafted classical autonomy stack and a human player.
   A complete list of hyperparameters is available in Appendix~\ref{app:hparams}.

\begin{itemize}
  \item \textbf{DreamerV3} As a model-based baseline with learned memory we use DreamerV3~\cite{dreamerv3} (official JAX implementation, 12M-parameter preset), whose recurrent latent state is the policy's only memory.
    Since the 2DSDS observation is a vector rather than an image, the encoder and decoder reduce to their MLP branches.
    We also test the MemoryMaze~\cite{memorymaze} memory variants (truncated BPTT with carried state, longer sequences) in \autoref{sec:adaptive_experiment}.
  \item \textbf{PPO} As a model-free, memoryless baseline we use PPO~\cite{ppo} from Stable-Baselines3~\cite{stable-baselines3} with the default two-layer MLP actor and critic on the flattened observation, without frame stacking or recurrence.
  \item \textbf{Frontier Explorer} For the multi-object search worlds, we implemented a standard frontier-based exploration \cite{paper_frontier_grandpa} stack in ROS2. 
    It builds a 2D occupancy grid using the pose and raycast data, greedily selects the nearest frontier cell on that grid, plans a path to it, follows the path, and selects the next frontier.
    The stack also switches to navigating towards a reward object whenever it sees one, then switches back to exploration.
    More implementation details are in \autoref{sec:frontier-imple}.
  \item \textbf{Human Player} - The data was collected from one person and is meant to serve as an estimated upper bound for evaluating performance of the other approaches. The player first played a few episodes of each scenario to get an understanding of each of them, and then was evaluated on before unseen seeds of the scenarios.
\end{itemize}

\subsection{Testing Effects of Topology, Dynamics and Task Complexity}
\label{sec:8worlds_experiment}
\begin{figure}[htb]
\centering
  \includegraphics[width=1.0\textwidth]{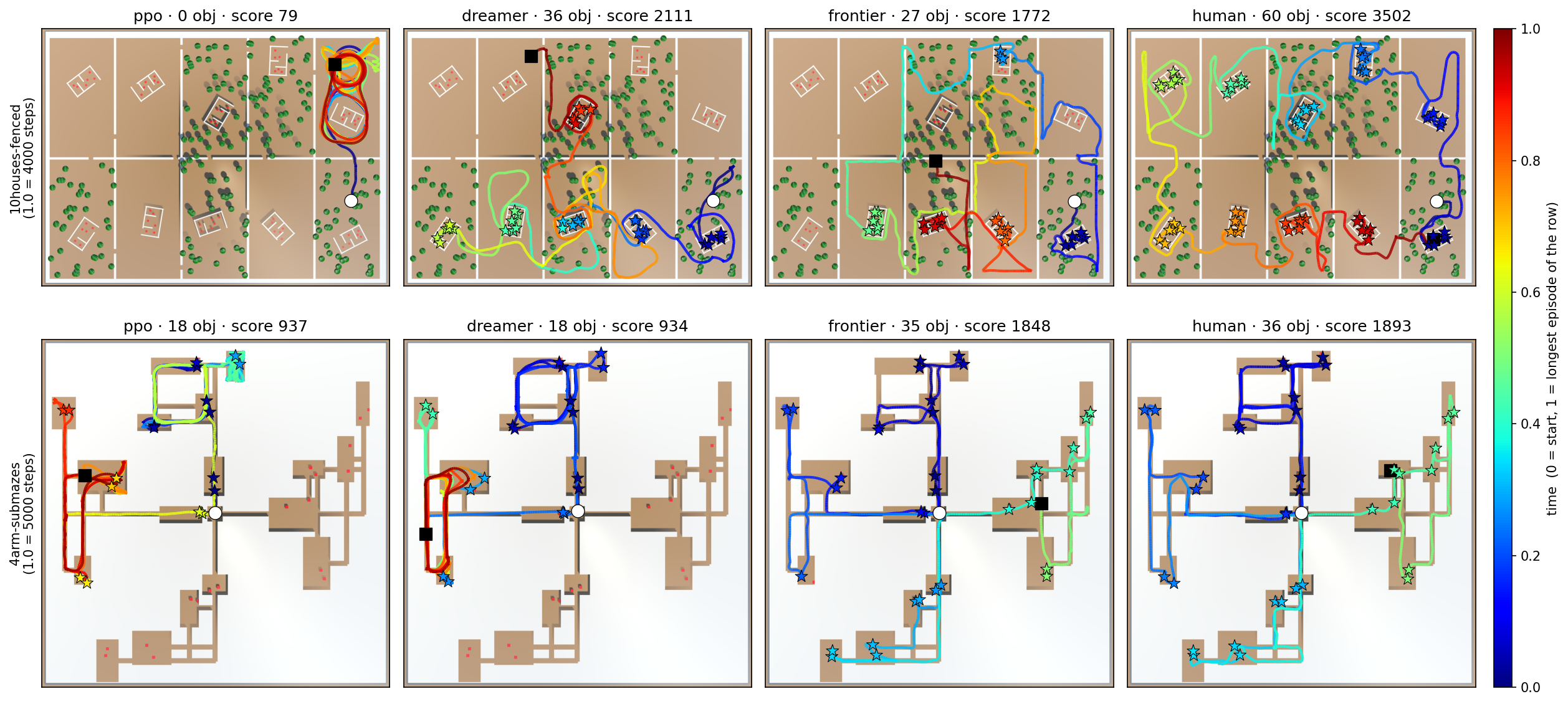}
  \caption{Visualization of eval trajectories on the \scenario{10houses-fenced} and \scenario{4arm-submazes} topologies. Color of trajectory indicates step index in episode, white circle = start, white square = end, stars = reward object pickups.}
\label{fig:trajs_topology}
\end{figure}

We evaluate the learned, engineered, and human baselines on 8 representative world configurations generated by \bench, with example trajectories from eval runs shown in \autoref{fig:trajs_topology} and in \autoref{sec:trajectories}.
We separate the scenarios into simple multi-object search in static environments, multi-object search in dynamic environments, and the well-alternation task.
The frontier exploration baseline is not evaluated on the alternation task as the behavior module would require significant reengineering for this specific task (both in the behavior and in having to do SLAM since we do not provide the agent's position as input in those experiments).
We originally ran the well-alternation task with the position-relative-to-spawn observation given to the agent, which allowed Dreamer to discover a trivial solution in the sense of "if well-direction-signal is not available -> navigate to position (0,0) and dodge obstacles reactively".
For this reason, the well-alternation runs shown in \autoref{tab:results} do not use position observations.
The human does not receive position information in any of the experiments, but still can still outperform baselines in nearly all runs.

\begin{figure}[htb]
\centering
\includegraphics[width=1.0\textwidth]{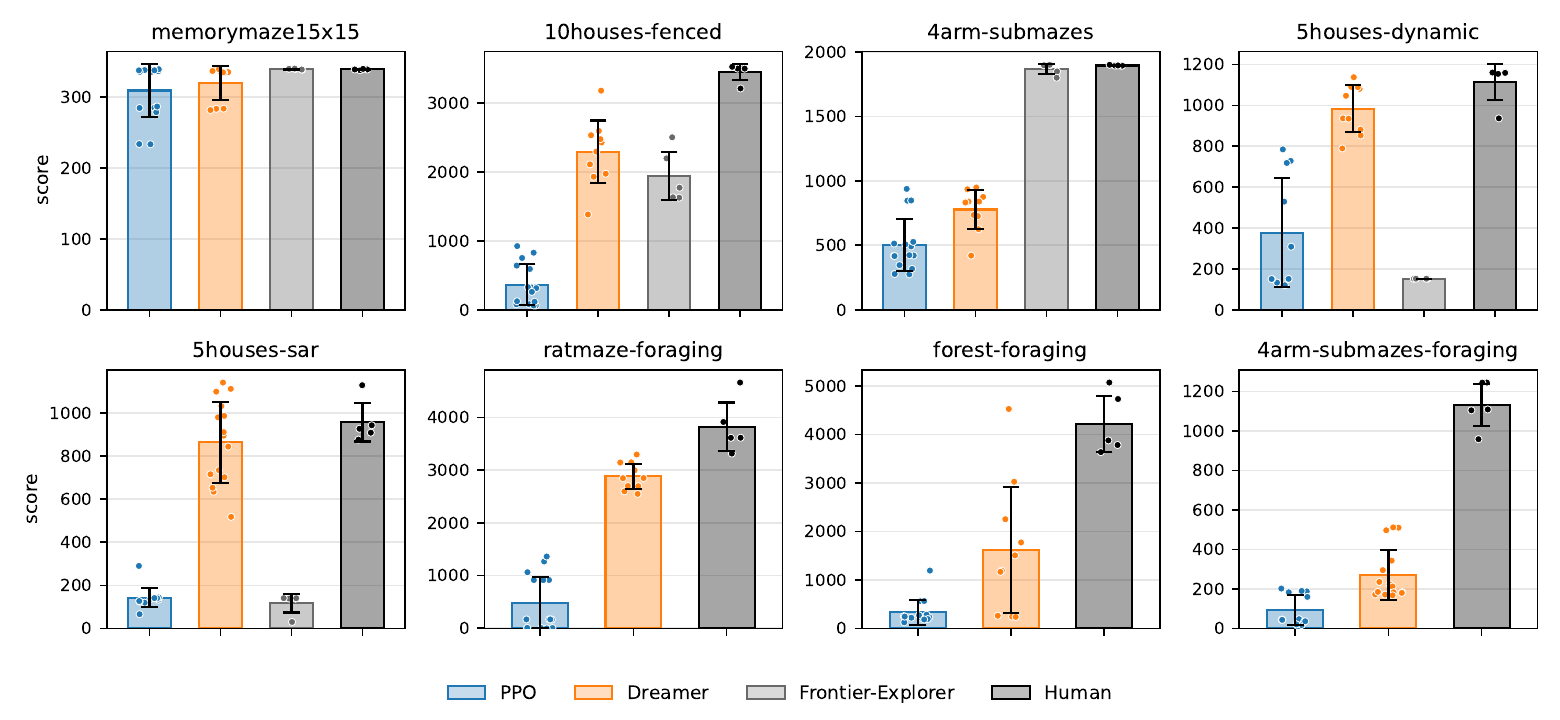}
  \caption{Bar charts of all the eval runs for each of the 8 scenarios described in \autoref{sec:8worlds_experiment}}
\label{fig:boxplots_full}
\end{figure}

We trained the RL baselines for a different number of training steps per scenario since the time to convergence varied drastically across scenarios (e.g. 1.5M steps in \scenario{memorymaze15x15} vs 60M steps in \scenario{forest-foraging}), with 3 seeds per method.
Episode lengths varied from 1000 to 5000 across scenarios.
We trained using 4 parallel envs per training run and A100 GPUs for Dreamer (PPO on CPU).
Full training curves are shown in \autoref{fig:training_curves_full} and detailed world descriptions in \autoref{sec:scenario_desc}.
To obtain the resulting data in \autoref{tab:results}, we evaluated all training seeds from each fully trained baseline on a set of 5 separate randomly generated worlds from each scenario.
Results of the eval runs are shown in \autoref{fig:boxplots_full} and \autoref{tab:results}, with object and coverage metrics in \autoref{tab:results_appendix}.

Our main general observation is that in essentially all the settings we tested, in both the coverage task and the well-alternation task, PPO and Dreamer converged to random-walk-like behaviors which often inefficiently revisited areas multiple times.
We found that these learned strategies can reach relatively high performance in surprisingly large scales and topologies
, especially if the environment is bounded and highly interconnected, but they fail on more complex topologies in which, in general, a wrong turn can lead to very disadvantageous state (e.g. getting lost in one of the sub-mazes of the \scenario{4arm-submazes} topology).

One example of an easy scenario in the multi-object search task is the largest variant of the MemoryMaze topology which we reproduced from \cite{memorymaze}, in which all our tested baselines reach a near perfect score, suggesting that coverage-oriented tasks in indoor-scale environments do not require any mapping or planning at all (which might not be that surprising given the popularity of even the cheaper, non-mapping robot vacuums).
We then tested whether a larger environment with more 'bottlenecks' could be a better test of map-based reasoning and found that indeed, on the \scenario{10houses-fenced} scenario, the random walk-like policy learned by the RL baselines struggled on account of often revisiting the fenced plots.
This scenario also serves as a failure case of the frontier explorer, since uniform coverage is not feasible in the alloted episode length, whereas a human player is able to understand that the rewards are always within the houses and can exploit this to find all the reward objects in time. 
In the \scenario{4arm-submazes} topology, both PPO and Dreamer struggled significantly and did not find a policy that would search the whole environment, 
while it was fairly easy for both a human and the frontier explorer to cover the maze fully.

To test the effect of environment dynamics, we simulated a setup of 5 randomly scattered houses in a bounded environment (which was a trivial topology for all baselines to fully cover when no dynamics were present) and tested first a setup where the agent has to push through rubble to get inside houses in the \scenario{5houses-dynamic} world.
In \scenario{5houses-sar}, we kept the same topology but made the dynamics harder by making just one of the rubble-walls movable (one was always too heavy to push through although they looked the same) and by adding sensor-blocking smoke surrounding the houses and chaotically walking objects.
As expected, both scenarios broke the frontier exploration approach, since it assumes a static environment.
Additionally, PPO struggled to learn in the harder-dynamics scenario, but Dreamer managed to reach near-human performance on both.

To test the effect of task complexity, we evaluated multiple worlds with the well-alternation task.
We observed that in the \scenario{ratmaze-foraging} scenario (a near 1:1 recreation of the experiment done in \cite{widloskiFlexibleReroutingHippocampal2022}), Dreamer achieved approx. 75\% of human performance, which can be attributed to the fact that the environment is small, bounded and well-connected and thus Dreamer was able to reach this score even with a relatively simple policy that revisited wells often.
For this reason, we then ran experiments in an open-ended environment in \scenario{forest-foraging} (where trees continue to spawn infinitely even if the agent leaves the 'content area' with the landmark objects and wells). 
Surprisingly, we found that Dreamer was able to learn a policy that mostly stays in the 'content area', but its overall performance was lower than human and had a very high per-world variance.
We also tested the same topology as in \scenario{4arm-submazes}, but with the well-alternation task, and found that here the RL baselines struggle extremely compared to human performance, indicating that it is the topology itself that is difficult to handle without map-based reasoning.

\begin{table}[!htb]
\centering
\small
\setlength{\tabcolsep}{4pt}
\renewcommand{\arraystretch}{1.15}
\providecommand{\cellEval}[1]{#1}
\providecommand{\cellTrain}[1]{{\color{blue}#1}}
\providecommand{\cellPartial}[1]{{\color{orange}#1}}
\providecommand{\cellMissing}{\textcolor{red}{--}}
\providecommand{\cellNA}{--}
\begin{tabular}{l l | l | l | l}
\toprule
\textbf{World} & \multicolumn{1}{c|}{\textbf{PPO}} & \multicolumn{1}{c|}{\textbf{Dreamer}} & \multicolumn{1}{c|}{\textbf{Frontier-Explorer}} & \multicolumn{1}{c}{\textbf{Human}} \\
\midrule
memorymaze15x15 & $309_{\cellEval{\pm 39}}$ & $320_{\cellEval{\pm 26}}$ & $\mathbf{339}_{\cellEval{\pm 1}}$ & $\mathbf{339}_{\cellEval{\pm 1}}$ \\
10houses-fenced & $366_{\cellEval{\pm 303}}$ & $2292_{\cellEval{\pm 481}}$ & $1947_{\cellEval{\pm 389}}$ & $\mathbf{3452}_{\cellEval{\pm 134}}$ \\
4arm-submazes & $504_{\cellEval{\pm 210}}$ & $777_{\cellEval{\pm 160}}$ & $1868_{\cellEval{\pm 44}}$ & $\mathbf{1895}_{\cellEval{\pm 3}}$ \\
\midrule
5houses-dynamic & $377_{\cellEval{\pm 282}}$ & $984_{\cellEval{\pm 120}}$ & $152_{\cellEval{\pm 0}}$ & $\mathbf{1114}_{\cellEval{\pm 100}}$ \\
5houses-sar & $142_{\cellEval{\pm 45}}$ & $864_{\cellEval{\pm 196}}$ & $118_{\cellEval{\pm 49}}$ & $\mathbf{957}_{\cellEval{\pm 99}}$ \\
\midrule
ratmaze-foraging & $470_{\cellEval{\pm 525}}$ & $2883_{\cellEval{\pm 256}}$ & \cellNA & $\mathbf{3827}_{\cellEval{\pm 515}}$ \\
forest-foraging & $335_{\cellEval{\pm 269}}$ & $1619_{\cellEval{\pm 1370}}$ & \cellNA & $\mathbf{4220}_{\cellEval{\pm 639}}$ \\
4arm-submazes-foraging & $94_{\cellEval{\pm 79}}$ & $272_{\cellEval{\pm 131}}$ & \cellNA & $\mathbf{1133}_{\cellEval{\pm 119}}$ \\
\bottomrule
\end{tabular}

  \caption{Benchmark results on the 8 world configurations. 
  All numbers show mean per-episode return and its standard deviation, computed from 5 eval seeds per scenario, averaged across training seeds}
\label{tab:results}
\end{table}

\subsection{Difficulty Curriculum Experiment}
\label{sec:adaptive_experiment}
Based on the low performance of learned baselines in \scenario{4arm-submazes-foraging} (not only the foraging task but also the multi-object search), we investigated whether Dreamer could learn better if the scale and complexity of the environment is smoothly increased as the agent's performance rises.
Similar to \cite{Portelas2020AutomaticCL}, we use an adaptive difficulty curriculum where we define a single scalar "difficulty" (0-1) which controls the scale of the environment, mainly the length of the maze center-arms and the number of rooms in each submaze, visualized in \autoref{fig:allworlds_adaptive}.
The curriculum defines a 'success' as finding 8 objects or more during one episode, which corresponds to navigating to a random well, back to the home well and to one more random well.
On a successful episode, the difficulty rises by 0.025, otherwise it decreases by 0.025.
\begin{wrapfigure}{r}{0.5\textwidth}
  \centering
  \includegraphics[width=0.48\textwidth]{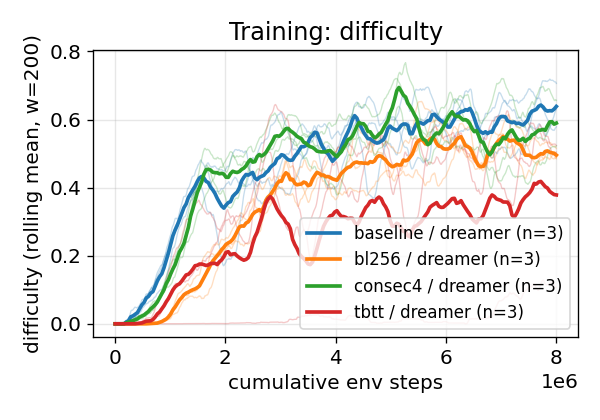}
  \caption{Adaptive difficulty curves for the curriculum experiment described in \autoref{sec:adaptive_experiment}.}
  \label{fig:adaptive_difficulty_curves}
\end{wrapfigure}
Since we run multiple environments in parallel (4 in our experiments), we track one difficulty value for all parallel envs, and we apply the step up or down whenever any one env finishes an episode. 
We additionally try to aid Dreamer's learning by increasing memory-related parameters. 
The bl256 variant lengthens the backpropagation window to 256 steps, while consec4 keeps the
  48-step gradient window but chains 4 consecutive windows with the recurrent state carried across them.
We also implement the trunacted backpropagation through time (TBTT) used in \cite{memorymaze} where the authors note that it increased performance in a memory task in mazes.
\autoref{fig:adaptive_difficulty_curves} shows how the adaptive difficulty changes during training for each variant.
Each variant converged betweeen 0.4 and 0.6 difficulty (tbtt and consec4 slower than bl256 and baseline), and neither was able to reach the full difficulty of \scenario{4arm-submazes-foraging}.


\section{Conclusion}
\label{sec:conclusion}

We have presented \bench, which we designed as a cross-disciplinary platform 
for evaluating both learned and engineered approaches on tasks that require map-based reasoning.
Through quantitative comparison of RL baselines, human performance, and the performance of a classical map-based autonomy stack on 8 representative scenarios, we show that environment topology is the main factor exposing the limitations of RL baselines in both the coverage and well-alternation tasks, and our results suggest that the presented well-alternation task is a stronger test of map-based reasoning than pure search or coverage since it requires the agent both to explore and to return to known locations. 
Furthermore, we performed an adaptive-curriculum expeirment in the complex maze scenario, smoothly scaling the maze size and complexity and found that neither this curriculum nor memory-related parameter tuning 
helped Dreamer to solve the task at full difficulty.
The comparatively high performance of the simple autonomy stack and the human suggests that map-based reasoning is a key limiting factor of RL baselines in these scenarios,
making \bench well-suited for future development of end-to-end learnable map-based reasoning architectures. 
\clearpage

\subsection*{AI Use Statement}
In this work, we used generative AI agents during the implementation of the majority of the \bench simulator (although the core of the architecture was written by hand), the training scheduler and evaluation tooling, the analysis and plotting scripts, the TBTT training variant, the frontier-explorer method, and code documentation. 
We have not used generative AI tools for generating synthetic datasets, developing theoretical models or conceptual frameworks, proposing or refining hypotheses, designing or providing feedback on research methodology, assisting with translation, cleaning or reformatting of datasets, supporting qualitative or thematic data analysis or interpreting results.
The categories of 
'formulate mathematical claims, provide critical ingredients for proving mathematical claims, assist in the writing of proofs,' 
 are not applicable to this work as we have no mathematical claims or proofs.
Additionally, we used generative AI tools for double-checking whether the parameters written in the manuscript reflect those in the codebase, searching for potential missed related works to cite, and for improving the readability of the paper by fixing typos, grammar and by chatting about the hard-to-read sections of the draft.
We have reviewed all AI-assisted work.
The majority of the code for which we used AI assistants was easy to check due to the nature of the work, for example the world generation code was checked by generating multiple worlds and seeing if the code did what the authors intended.
The ROS2-based frontier-exploration node's behavior was checked by visualizations both in the simulator and through standard RVIZ visualizations of the occupancy map, planned path etc. 
The evaluations were verified by running the evaluations inside the Unity editor (i.e. not in headless mode) and observing whether the analyzed data (the trajectory, score etc.) matched the behavior of the agent in the simulation.
The training-related code written with the help of AI agents is mostly wrappers, environment adapters and orchestration of runs and logging, with the core training logic unchanged from the Stable-Baselines3 and the Dreamer's original JAX implementation.
We verified the TBTT implementation by logging whether the recurrent state is carried correctly.
The training data was visualized through WeightsAndBiases.
We take responsibility for the final content of this work,
including text, claims or artifacts produced with the aid of generative AI.

\subsection*{Ethics Statement}
This work presents a simulation environment and benchmark.
It involves no personally identifiable data and no experiments on animals. The only human data are the gameplay episodes of a single volunteer participant collected for the human reference baseline, with informed consent and without any personal information being recorded.
Some environments ForagerSim are inspired by search-and-rescue scenarios and simulate only abstract agents, obstacles and reward objects, it does not model or simulate any harm to people.
We release it to support research in robotics, reinforcement learning and cognitive science and see no immediate risks beyond those common to the field and the related simulation environments.
We reuse the experimental protocol of \cite{widloskiFlexibleReroutingHippocampal2022} in simulation only, no animal data are used.

\subsection*{Reproducibility Statement}
All experiments in this paper use the ForagerSim simulator, which we share as a Linux build together with the Unity source code, Python wrapper, the Gym and ROS 2 interfaces, the eight world, agent and task configuration files used in Section 4, and the training and evaluation scripts, in the supplementary material.
They contain a README file with instructions for running the simulation both in a human-control mode and for running the training, evaluation and analysis as done in this paper.
World generation is deterministic given the seeds listed in the configuration files (saved in a JSON alongside the eval data).
Hyperparameters for all RL baselines, including every deviation from library defaults, are listed in \autoref{app:hparams}.
The frontier-explorer implementation is described in \autoref{sec:frontier-imple}.
The evaluation protocol (number of training seeds, evaluation worlds and episode lengths) is stated in \autoref{sec:8worlds_experiment} and \autoref{sec:scenario_desc}.
The recorded trajectories from all baselines including the human reference are shared in the supplementary materials.

\subsection*{Acknowledgments}
This work was supported by the European Union under the project Robotics and advanced industrial production (reg. no. CZ.02.01.01/00/22\_008/0004590), by the Czech Science Foundation (project no. GA21-31000S), and by CTU grant no. SGS26/077/OHK3/1T/13.
T.~Musil gratefully acknowledges financial support for this research by the
Fulbright Foreign Student Program (Visiting Student Researcher), which is
sponsored by the U.S. Department of State and the J. William Fulbright
Commission in the Czech Republic. Its contents are solely the responsibility
of the authors and do not necessarily represent the official views of the
Fulbright Program, the Government of the United States, or the J. William
Fulbright Commission in the Czech Republic.

\bibliographystyle{iclr2027_conference}
\bibliography{main}

@article{aronovMappingNonspatialDimension2017,
  title = {Mapping of a Non-Spatial Dimension by the Hippocampal--Entorhinal Circuit},
  author = {Aronov, Dmitriy and Nevers, Rhino and Tank, David W.},
  year = 2017,
  month = mar,
  journal = {Nature},
  volume = {543},
  number = {7647},
  pages = {719--722},
  publisher = {Nature Publishing Group},
  langid = {english}
}

@article{widloskiFlexibleReroutingHippocampal2022,
  title = {Flexible Rerouting of Hippocampal Replay Sequences around Changing Barriers in the Absence of Global Place Field Remapping},
  author = {Widloski, John and Foster, David J.},
  year = 2022,
  month = may,
  journal = {Neuron},
  volume = {110},
  number = {9},
  pages = {1547--1558.e8},
  publisher = {Elsevier},
  langid = {english}
}

@article{baraducSchemaCellsMacaque2019,
  title = {Schema Cells in the Macaque Hippocampus},
  author = {Baraduc, Pierre and Duhamel, Jean-Ren{\'e} and Wirth, Sylvia},
  year = 2019,
  month = feb,
  journal = {Science},
  volume = {363},
  number = {6427},
  pages = {635--639},
  publisher = {American Association for the Advancement of Science},
  langid = {english}
}

@article{tolmanCognitiveMapsRats1948,
  title = {Cognitive Maps in Rats and Men},
  author = {Tolman, Edward C.},
  year = 1948,
  journal = {Psychological Review},
  volume = {55},
  number = {4},
  pages = {189--208},
  publisher = {American Psychological Association},
  langid = {english}
}

@article{Epstein2017TheCM,
  title={The cognitive map in humans: Spatial navigation and beyond},
  author={Russell A. Epstein and Eva Zita Patai and Joshua B. Julian and Hugo J. Spiers},
  journal={Nature Neuroscience},
  year={2017},
  volume={20},
  number={11},
  pages={1504--1513},
  doi={10.1038/nn.4656},
}

@article{gevasagivSpatialCognitionBats2015,
  title = {Spatial Cognition in Bats and Rats: From Sensory Acquisition to Multiscale Maps and Navigation},
  shorttitle = {Spatial Cognition in Bats and Rats},
  author = {{Geva-Sagiv}, Maya and Las, Liora and Yovel, Yossi and Ulanovsky, Nachum},
  year = 2015,
  month = feb,
  journal = {Nature Reviews Neuroscience},
  volume = {16},
  number = {2},
  pages = {94--108},
  publisher = {Nature Publishing Group},
  langid = {english}
}

@article{wolves_navigation,
  title={Spatial Memory Drives Foraging Strategies of Wolves, but in Highly Individual Ways},
  author={Eliezer Gurarie and Chloe Bracis and Angelina Brilliantova and Ilpo Kojola and Johanna Suutarinen and Otso Ovaskainen and Sriya Potluri and William F. Fagan},
  journal={Frontiers in Ecology and Evolution},
  year={2022},
  volume={10},
  pages={768478},
  doi={10.3389/fevo.2022.768478},
}

@article{
kidnapped_bats,
author = {Aya Goldshtein  and Xing Chen  and Eran Amichai  and Arjan Boonman  and Lee Harten  and Omer Yinon  and Yotam Orchan  and Ran Nathan  and Sivan Toledo  and Iain D. Couzin  and Yossi Yovel },
title = {Acoustic cognitive map–based navigation in echolocating bats},
journal = {Science},
volume = {386},
number = {6721},
pages = {561-567},
year = {2024},
doi = {10.1126/science.adn6269},
}

@article{animal_ai_bench,
  title={The Animal-AI Environment: A virtual laboratory for comparative cognition and artificial intelligence research},
  author={Konstantinos Voudouris and Ibrahim Alhas and Wout Schellaert and Matteo G. Mecattaf and Ben Slater and Matthew Crosby and Joel Holmes and John Burden and Niharika Chaubey and Niall Donnelly and Matishalin Patel and Marta Halina and Jos{\'e} Hern{\'a}ndez-Orallo and Lucy G. Cheke},
  journal={Behavior Research Methods},
  year={2025},
  volume={57},
  number={4},
  pages={107},
  doi={10.3758/s13428-025-02616-3},
}

@inproceedings{minerl,
  title={MineRL: A Large-Scale Dataset of Minecraft Demonstrations},
  author={William H. Guss and Brandon Houghton and Nicholay Topin and Phillip Wang and Cayden R. Codel and Manuela M. Veloso and Ruslan Salakhutdinov},
  booktitle={International Joint Conference on Artificial Intelligence},
  pages={2442--2448},
  year={2019},
  doi={10.24963/ijcai.2019/339},
}

@inproceedings{minedojo,
  title={MineDojo: Building Open-Ended Embodied Agents with Internet-Scale Knowledge},
  author={Linxi Fan and Guanzhi Wang and Yunfan Jiang and Ajay Mandlekar and Yuncong Yang and Haoyi Zhu and Andrew Tang and De-An Huang and Yuke Zhu and Anima Anandkumar},
  booktitle={Advances in Neural Information Processing Systems (NeurIPS), Datasets and Benchmarks Track},
  year={2022},
}

@inproceedings{memorymaze,
  title={Evaluating Long-Term Memory in 3D Mazes},
  author={Jurgis Pa\v{s}ukonis and Timothy P. Lillicrap and Danijar Hafner},
  booktitle={International Conference on Learning Representations (ICLR)},
  year={2023},
}

@inproceedings{balrog,
  title={{BALROG}: Benchmarking Agentic {LLM} and {VLM} Reasoning On Games},
  author={Davide Paglieri and Bart{\l}omiej Cupia{\l} and Samuel Coward and Ulyana Piterbarg and Maciej Wo{\l}czyk and Akbir Khan and Eduardo Pignatelli and {\L}ukasz Kuci{\'n}ski and Lerrel Pinto and Rob Fergus and Jakob N. Foerster and Jack Parker-Holder and Tim Rockt{\"a}schel},
  booktitle={International Conference on Learning Representations (ICLR)},
  year={2025},
}

@inproceedings{nethack,
  title={{The NetHack Learning Environment}},
  author={Heinrich K{\"u}ttler and Nantas Nardelli and Alexander H. Miller and Roberta Raileanu and Marco Selvatici and Edward Grefenstette and Tim Rockt{\"a}schel},
  booktitle={Advances in Neural Information Processing Systems (NeurIPS)},
  volume={33},
  year={2020},
}

@inproceedings{minihack,
  title={{MiniHack the Planet: A Sandbox for Open-Ended Reinforcement Learning Research}},
  author={Mikayel Samvelyan and Robert Kirk and Vitaly Kurin and Jack Parker-Holder and Minqi Jiang and Eric Hambro and Fabio Petroni and Heinrich K{\"u}ttler and Edward Grefenstette and Tim Rockt{\"a}schel},
  booktitle={Advances in Neural Information Processing Systems (NeurIPS)},
  year={2021},
}

@inproceedings{crafter,
  title={{Benchmarking the Spectrum of Agent Capabilities}},
  author={Danijar Hafner},
  booktitle={International Conference on Learning Representations (ICLR)},
  year={2022},
}

@inproceedings{avalon,
  title={{Avalon: A Benchmark for RL Generalization Using Procedurally Generated Worlds}},
  author={Joshua Albrecht and Abraham J. Fetterman and Bryden Fogelman and Ellie Kitanidis and Bartosz Wr{\'o}blewski and Nicole Seo and Michael Rosenthal and Maksis Knutins and Zachary Polizzi and James B. Simon and Kanjun Qiu},
  booktitle={Advances in Neural Information Processing Systems (NeurIPS), Datasets and Benchmarks Track},
  year={2022},
}

@article{ppo,
  title={{Proximal Policy Optimization Algorithms}},
  author={John Schulman and Filip Wolski and Prafulla Dhariwal and Alec Radford and Oleg Klimov},
  journal={arXiv preprint arXiv:1707.06347},
  year={2017},
}

@inproceedings{gymnasium,
  title={{Gymnasium: A Standard Interface for Reinforcement Learning Environments}},
  author={Mark Towers and Ariel Kwiatkowski and Jordan Terry and John Balis and Gianluca De Cola and Tristan Deleu and others},
  booktitle={Advances in Neural Information Processing Systems (NeurIPS), Datasets and Benchmarks Track},
  year={2025},
}

@inproceedings{raven,
  title={{RAVEN}: Resilient Aerial Navigation via Open-Set Semantic Memory and Behavior Adaptation},
  author={Seungchan Kim and Omar Alama and Dmytro Kurdydyk and John Keller and Nikhil Keetha and Wenshan Wang and Yonatan Bisk and Sebastian Scherer},
  booktitle={IEEE International Conference on Robotics and Automation (ICRA)},
  year={2026},
  note={arXiv:2509.23563},
}

@INPROCEEDINGS{paper_frontier_grandpa,
  author={Yamauchi, B.},
  booktitle={Proceedings 1997 IEEE International Symposium on Computational Intelligence in Robotics and Automation CIRA'97},
  title={{A Frontier-Based Approach for Autonomous Exploration}},
  year={1997},
  pages={146-151},
  doi={10.1109/CIRA.1997.613851},
}

@article{darpa_cerberus_wins,
  title={{CERBERUS in the DARPA Subterranean Challenge}},
  author={Marco Tranzatto and Takahiro Miki and Mihir Dharmadhikari and Lukas Bernreiter and Mihir Kulkarni and Frank Mascarich and Olov Andersson and Shehryar Khattak and Marco Hutter and Roland Y. Siegwart and Kostas Alexis},
  journal={Science Robotics},
  year={2022},
  volume={7},
  number={66},
  pages={eabp9742},
  doi={10.1126/scirobotics.abp9742},
}

@article{csiro_darpa,
  title={{Heterogeneous Ground and Air Platforms, Homogeneous Sensing: Team CSIRO Data61's Approach to the DARPA Subterranean Challenge}},
  author={Nicolas Hudson and Fletcher Talbot and Mark Cox and Jason Williams and Thomas Hines and Alex Pitt and Brett Wood and Dennis Frousheger and Katrina Lo Surdo and Thomas Molnar and Ryan Steindl and Matt Wildie and Inkyu Sa and Navinda Kottege and Kazys Stepanas and Emili Hern{\'a}ndez and Gavin Catt and William Docherty and Brendan Tidd and Benjamin Tam and Simon Murrell and Mitchell Bessell and Lauren Hanson and Lachlan Tychsen-Smith and Hajime Suzuki and Leslie Overs and Farid Kendoul and Glenn Wagner and Duncan Palmer and Peter Milani and Matthew O'Brien and Shu Jiang and Shengkang Chen and Ronald C. Arkin},
  journal={Field Robotics},
  year={2022},
  volume={2},
  pages={595--636},
  doi={10.55417/fr.2022021},
}

@article{beneath,
  title={{UAVs Beneath the Surface: Cooperative Autonomy for Subterranean Search and Rescue in DARPA SubT}},
  author={Matěj Petrlík and Pavel Petráček and Vít Krátký and Tomáš Musil and Yurii Stasinchuk and Matouš Vrba and Tomáš Báča and Daniel Heřt and Martin Pecka and Tomáš Svoboda and Martin Saska},
  journal={IEEE Transactions on Field Robotics},
  year={2025},
  volume={2},
  pages={643--689},
  doi={10.1109/TFR.2024.3492160},
}

@article{dreamerv3,
  title={Mastering diverse domains through world models},
  author={Hafner, Danijar and Pasukonis, Jurgis and Ba, Jimmy and Lillicrap, Timothy},
  journal={Nature},
  year={2025},
  volume={640},
  number={8059},
  pages={647--653},
  doi={10.1038/s41586-025-08744-2},
}

@inproceedings{popgym,
  title={{POPGym}: Benchmarking Partially Observable Reinforcement Learning},
  author={Morad, Steven and Kortvelesy, Ryan and Bettini, Matteo and Liwicki, Stephan and Prorok, Amanda},
  booktitle={International Conference on Learning Representations (ICLR)},
  year={2023},
}

@inproceedings{procthor,
  title={{ProcTHOR}: Large-Scale Embodied {AI} Using Procedural Generation},
  author={Deitke, Matt and VanderBilt, Eli and Herrasti, Alvaro and Weihs, Luca and Salvador, Jordi and Ehsani, Kiana and Han, Winson and Kolve, Eric and Farhadi, Ali and Kembhavi, Aniruddha and Mottaghi, Roozbeh},
  booktitle={Advances in Neural Information Processing Systems (NeurIPS)},
  year={2022},
}

@inproceedings{segment_anything,
  title={Segment Anything},
  author={Kirillov, Alexander and Mintun, Eric and Ravi, Nikhila and Mao, Hanzi and Rolland, Chloe and Gustafson, Laura and Xiao, Tete and Whitehead, Spencer and Berg, Alexander C. and Lo, Wan-Yen and Doll{\'a}r, Piotr and Girshick, Ross},
  booktitle={IEEE/CVF International Conference on Computer Vision (ICCV)},
  pages={3992--4003},
  year={2023},
  doi={10.1109/ICCV51070.2023.00371},
}

@article{banino2018vector,
  title   = {Vector-based navigation using grid-like representations in artificial agents},
  author  = {Banino, Andrea and Barry, Caswell and Uria, Benigno and Blundell, Charles and
             Lillicrap, Timothy and Mirowski, Piotr and Pritzel, Alexander and Chadwick, Martin J. and
             Degris, Thomas and Modayil, Joseph and Wayne, Greg and Soyer, Hubert and Viola, Fabio and
             Zhang, Brian and Goroshin, Ross and Rabinowitz, Neil and Pascanu, Razvan and
             Beattie, Charlie and Petersen, Stig and Sadik, Amir and Gaffney, Stephen and King, Helen and
             Kavukcuoglu, Koray and Hassabis, Demis and Hadsell, Raia and Kumaran, Dharshan},
  journal = {Nature},
  volume  = {557}, number = {7705}, pages = {429--433}, year = {2018},
  doi     = {10.1038/s41586-018-0102-6}
}

@article{yangSpatiallyenhancedRecurrentMemory2025,
  title = {Spatially-Enhanced Recurrent Memory for Long-Range Mapless Navigation via End-to-End Reinforcement Learning},
  author = {Yang, Fan and Frivik, Per and Hoeller, David and Wang, Chen and Cadena, Cesar and Hutter, Marco},
  year = 2026,
  journal = {The International Journal of Robotics Research},
  volume = {45},
  number = {11},
  pages = {1682--1706},
  doi = {10.1177/02783649251401926},
  publisher = {SAGE Publications Ltd STM},
  langid = {english}
}

@inproceedings{wijmansEmergenceMapsMemories2023,
  title        = {Emergence of Maps in the Memories of Blind Navigation Agents},
  author       = {Wijmans, Erik and Savva, Manolis and Essa, Irfan and Lee, Stefan and Morcos, Ari S. and Batra, Dhruv},
  year         = {2023},
  booktitle    = {The Eleventh International Conference on Learning Representations (ICLR 2023)},
  eprint       = {2301.13261},
  eprinttype   = {arXiv}
}

@article{stable-baselines3,
  title={Stable-Baselines3: Reliable Reinforcement Learning Implementations},
  author={Raffin, Antonin and Hill, Ashley and Gleave, Adam and Kanervisto, Anssi and Ernestus, Maximilian and Dormann, Noah},
  journal={Journal of Machine Learning Research},
  volume={22},
  number={268},
  pages={1--8},
  year={2021},
}

@inproceedings{Portelas2020AutomaticCL,
  title={Automatic Curriculum Learning For Deep {RL}: A Short Survey},
  author={R{\'e}my Portelas and C{\'e}dric Colas and Lilian Weng and Katja Hofmann and Pierre-Yves Oudeyer},
  booktitle={Proceedings of the Twenty-Ninth International Joint Conference on Artificial Intelligence (IJCAI)},
  pages={4819--4825},
  year={2020},
  doi={10.24963/ijcai.2020/671},
}

@inproceedings{explore_bench,
  title     = {{Explore-Bench}: Data Sets, Metrics and Evaluations for Frontier-based and Deep-reinforcement-learning-based Autonomous Exploration},
  author    = {Xu, Yuanfan and Yu, Jincheng and Tang, Jiahao and Qiu, Jiantao and Wang, Jian and Shen, Yuan and Wang, Yu and Yang, Huazhong},
  booktitle = {IEEE International Conference on Robotics and Automation (ICRA)},
  pages     = {6225--6231},
  year      = {2022},
  doi       = {10.1109/ICRA46639.2022.9812344}
}

@inproceedings{habitat,
  title     = {Habitat: A Platform for Embodied {AI} Research},
  author    = {Savva, Manolis and Kadian, Abhishek and Maksymets, Oleksandr and Zhao, Yili and Wijmans, Erik and Jain, Bhavana and Straub, Julian and Liu, Jia and Koltun, Vladlen and Malik, Jitendra and Parikh, Devi and Batra, Dhruv},
  booktitle = {IEEE/CVF International Conference on Computer Vision (ICCV)},
  pages     = {9338--9346},
  year      = {2019},
  doi       = {10.1109/ICCV.2019.00943}
}

@inproceedings{habitat3,
  title     = {Habitat 3.0: A Co-Habitat for Humans, Avatars, and Robots},
  author    = {Puig, Xavier and Undersander, Eric and Szot, Andrew and Cote, Mikael Dallaire and Yang, Tsung-Yen and Partsey, Ruslan and Desai, Ruta and Clegg, Alexander William and Hlavac, Michal and Min, So Yeon and Vondru{\v{s}}, Vladim{\'i}r and Gervet, Theophile and Berges, Vincent-Pierre and Turner, John M. and Maksymets, Oleksandr and Kira, Zsolt and Kalakrishnan, Mrinal and Malik, Jitendra and Chaplot, Devendra Singh and Jain, Unnat and Batra, Dhruv and Rai, Akshara and Mottaghi, Roozbeh},
  booktitle = {International Conference on Learning Representations (ICLR)},
  year      = {2024},
  note      = {arXiv:2310.13724}
}

@inproceedings{craftax,
  title     = {{Craftax}: A Lightning-Fast Benchmark for Open-Ended Reinforcement Learning},
  author    = {Matthews, Michael and Beukman, Michael and Ellis, Benjamin and Samvelyan, Mikayel and Jackson, Matthew Thomas and Coward, Samuel and Foerster, Jakob Nicolaus},
  booktitle = {Proceedings of the 41st International Conference on Machine Learning (ICML)},
  series    = {Proceedings of Machine Learning Research},
  volume    = {235},
  pages     = {35104--35137},
  year      = {2024},
  publisher = {PMLR}
}

@inproceedings{unrealzoo,
  title     = {{UnrealZoo}: Enriching Photo-realistic Virtual Worlds for Embodied {AI}},
  author    = {Zhong, Fangwei and Wu, Kui and Wang, Churan and Chen, Hao and Ci, Hai and Li, Zhoujun and Wang, Yizhou},
  booktitle = {IEEE/CVF International Conference on Computer Vision (ICCV)},
  year      = {2025},
  note      = {arXiv:2412.20977}
}

@inproceedings{airsim,
  title     = {{AirSim}: High-Fidelity Visual and Physical Simulation for Autonomous Vehicles},
  author    = {Shah, Shital and Dey, Debadeepta and Lovett, Chris and Kapoor, Ashish},
  booktitle = {Field and Service Robotics (FSR)},
  series    = {Springer Proceedings in Advanced Robotics},
  volume    = {5},
  pages     = {621--635},
  publisher = {Springer},
  year      = {2018},
  doi       = {10.1007/978-3-319-67361-5_40},
  note      = {arXiv:1705.05065}
}

@inproceedings{metaurban,
  title     = {{MetaUrban}: An Embodied {AI} Simulation Platform for Urban Micromobility},
  author    = {Wu, Wayne and He, Honglin and He, Jack and Wang, Yiran and Duan, Chenda and Liu, Zhizheng and Li, Quanyi and Zhou, Bolei},
  booktitle = {International Conference on Learning Representations (ICLR)},
  year      = {2025},
  note      = {arXiv:2407.08725}
}

@article{Pleines2023MemoryGT,
  title={Memory Gym: Towards Endless Tasks to Benchmark Memory Capabilities of Agents},
  author={Marco Pleines and Matthias Pallasch and Frank Zimmer and Mike Preuss},
  journal={Journal of Machine Learning Research},
  year={2025},
  volume={26},
  number={6},
  pages={1--40},
}

@book{gerritsenK9SearchRescue2014,
  title     = {K9 Search and Rescue: A Manual for Training the Natural Way},
  author    = {Gerritsen, Resi and Haak, Ruud},
  year      = {2014},
  edition   = {2},
  series    = {K9 Professional Training Series},
  publisher = {Dog Training Press (Brush Education)},
  address   = {Edmonton},
  isbn      = {9781550594478}
}


\appendix

\section{Detailed Scenario Descriptions}
\label{sec:scenario_desc}
\begin{figure}[htb]
\centering
\includegraphics[width=1.0\textwidth]{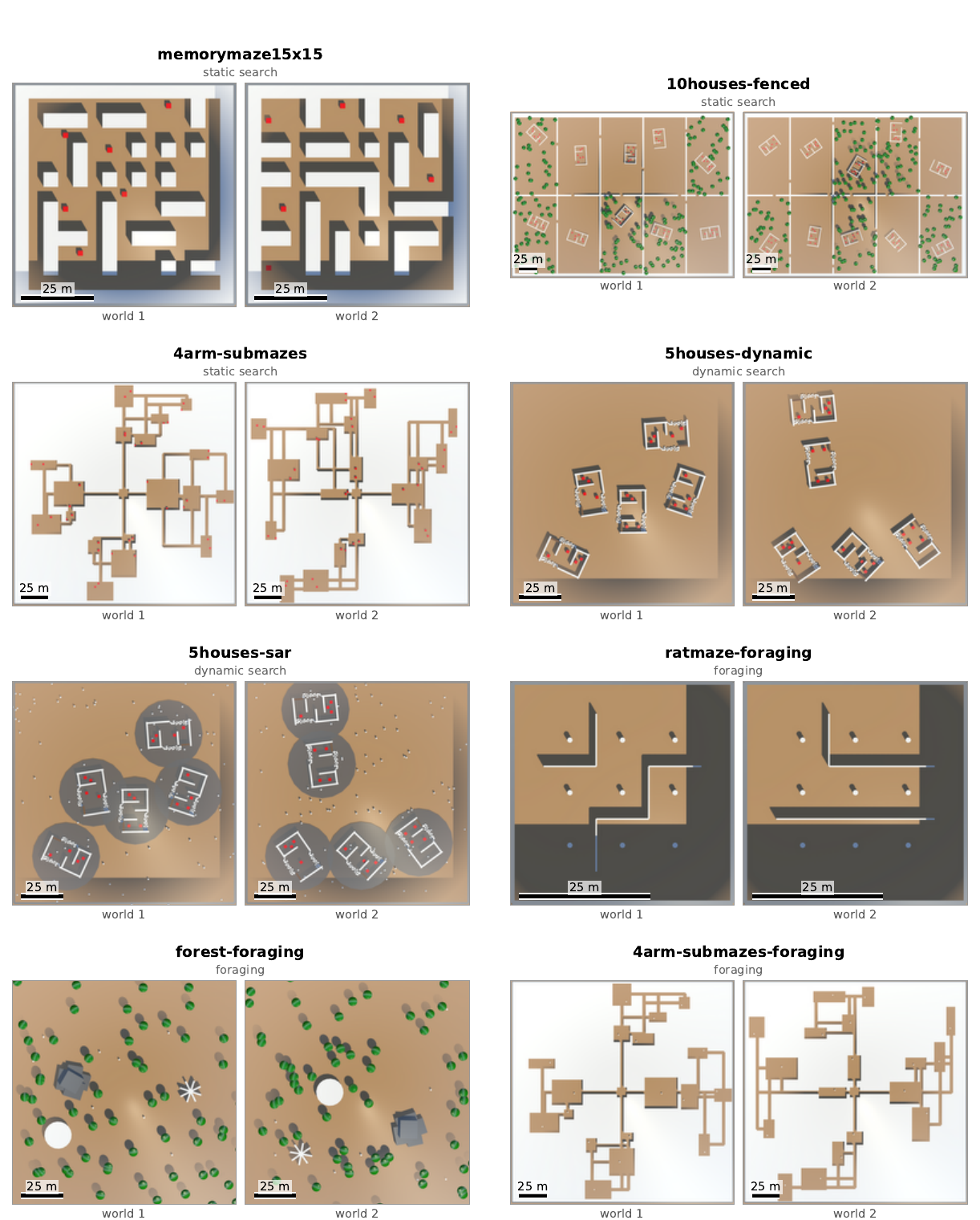}
  \caption{A visualization of 2 random seed for each of the 8 worlds used in the comparisons in \autoref{sec:8worlds_experiment}}
\label{fig:allworlds}
\end{figure}

\begin{figure}[htb]
\centering
\includegraphics[width=1.0\textwidth]{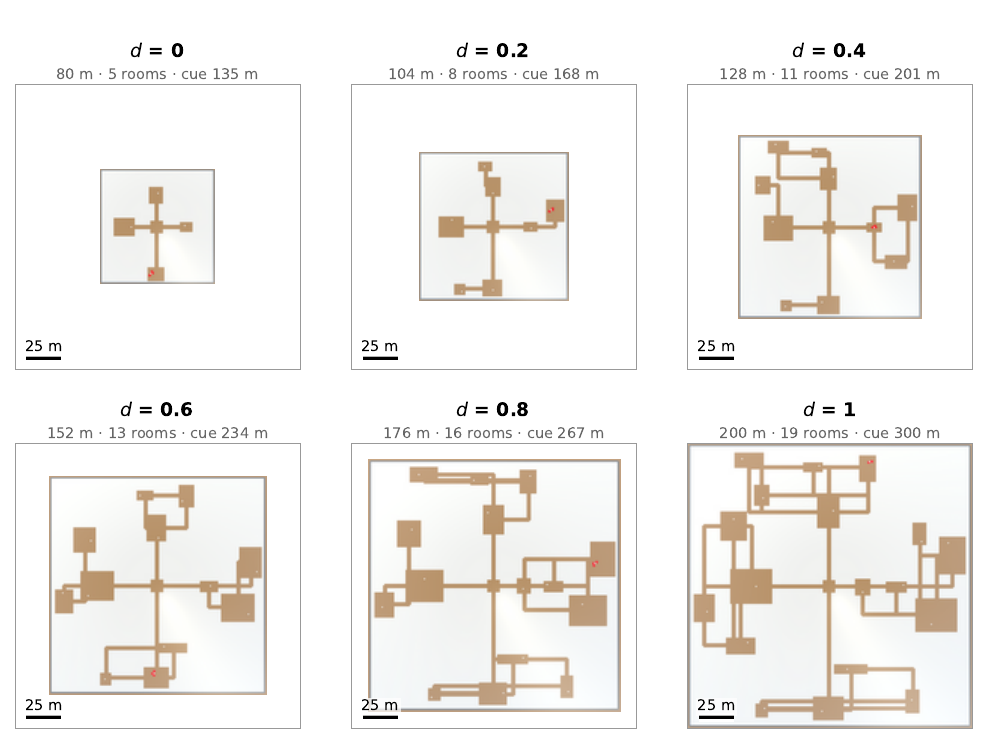}
  \caption{A visualization of adaptive difficulty scaling of world complexity on the example of the \scenario{4arm-submazes} world in the experiments described in \autoref{sec:adaptive_experiment}}
\label{fig:allworlds_adaptive}
\end{figure}

Here we describe the 8 presented scenarios in more detail and note how many steps are in each episode in any given scenario.
The episode lengths for each scenario were designed to be solvable by a human player.
Visualizations of 2 random seeds for each scenario are shown in in \autoref{fig:allworlds}, and visualizations of the effect of the difficulty in the curriculum experiment on the environmental scale and topology are visualized in \autoref{fig:allworlds_adaptive}.

\subsection{Static-Environment Multi-Object Search}

\paragraph{\scenario{memorymaze15x15}} (1000 steps, 75$\times$75\,m, 6 reward objects) is a recreation of the largest MemoryMaze environment from \cite{memorymaze}, implemented by rewriting the generation code used by the authors in DMLab into C\# inside the Unity project of \bench.
The environment is generated by placing a set of 6 rooms which contain 1 reward object each and 3 rooms that contain no reward (mimicking the placement of reward objects from the original, although in our task the agent can pick up the objects in any order), and the rooms are connected by corridors to be fully connected. 
The topologies remain the same as in the original, the only change is that we do not color the walls as in MemoryMaze.
\paragraph{\scenario{10houses-fenced}} (4000 steps, 298$\times$220\,m, 60 reward objects) is a scenario designed to emulate a section of a suburb in an outdoor search-and-rescue scenario, with 10 plots of land arranged in two rows, each with a single house and in 50\% cases with dense vegetation as well, where the plots are separated by a smooth wall which has 1 small hole at a random position between any 2 adjacent plots.
Each house contains 6 randomly placed reward objects, no objects can be found outside the houses.
Each plot is 56x104 meters, which, compared to the sensor range of 30 meters forces the agent to \textit{search} for both the house, the entrance to the house and for the hole between adjacent cells, which is even harder in the vegetation-dense cells since they sometimes block these objects.
\paragraph{\scenario{4arm-submazes}} (5000 steps, 200$\times$200\,m, 36 reward objects) demonstrates the multi-object search task in a maze topology which we designed specifically to break any random-walk strategies, which the RL baselines often converged to.
The maze topology consists of 1 central chamber, with 4 arms leading in orthogonal directions from it, and a submaze at the end of each arm. 
Each submaze is generated by randomly placing randomly shaped rectangular rooms and connecting them so that they are fully connected within the submaze, but only remain connected to the other submazes through the central chamber.
This way, policies that do not explicitly stop themselves from revisiting areas often spend a large portion of the episode wandering inside the submazes.

\subsection{Dynamic-Environment Multi-Object Search}
\paragraph{\scenario{5houses-dynamic}} (2000 steps, 130$\times$130\,m, 20 reward objects) is a multi-object search scenario designed to show the effects of simple world dynamics in a simple world topology where 5 house objects are randomly placed and rotated within a box arena.
Each house contains 4 randomly placed reward objects.
This is a topology with no bottlenecks and without the dynamics, both RL policies and the frontier baseline are able to search it fully.
The dynamics are simple in the sense that each house is fully enclosed with the exception of 2 walls which are converted into 'rubble walls'. 
However, even with these simple dynamics (motivated by a real real-world problem of a robot having to push a few objects out of its way to get somewhere) force the frontier exploration approach to never reach inside the houses and would require some scenario-specific engineering to force the agent to push through the boxes and to clear its map of the dynamic objects.
These rubble walls consist of approx. 15 movable boxes of variable sizes which block the view into the house and which have a different semantic class than obstacles and reward objects.
The agent can reach inside the house simply by pushing through the rubble, however it then has to maneuver around the rubble inside the house.
\paragraph{\scenario{5houses-sar}} (2000 steps, 130$\times$130\,m, 20 reward objects) is a harder variant of \scenario{5houses-dynamic}, where on each house, one wall has the mass of its rubble objects set extremely high so that they are impossible to move by the agent but it looks the same as the movable rubble wall.
In addition, smoke is present around each house and it degrades the sensing by simulating random lidar hits from smoke particles.
The smoke has the same semantic class as obstacles.
The environment also contains chaotically moving objects (implemented as simple capsules) out of which 50\% will always move out of the way of the agent and 50\% will not care about the agent, emulating people with different behaviors moving in the area (but fundamentally this aims to simulate any chaotically moving objects in the scenario).
The chaotic walkers have their own semantic class.

\subsection{Foraging (Well Alternation)}
All three foraging worlds use the same well scheduler and exactly one well is armed at any time.
A well dispenses when the agent stays within 3 m of it for 10 consecutive steps.
It then spawns 3 reward objects 1–1.7 m from the well, and the next well is armed after all of them are collected.
Well priming then alternate between random and home.
During random priming, a non-home well is drawn uniformly, excluding the previous random well, and it is cued.
The cue is an 8-sector egocentric signal, which for each angular bin (45 degrees) has a value ranging from 0-1 which decreases with the angle between the bin normal and the direction to the cue, and with distance to the cue.  
Its range covers the whole arena and it provides direction and distance information for the armed random well, but provides no information about home wells (all values are 0 during return to home).
Every well looks the same (however it has a different semantic class than other objects) regardless of whether it is a home or random well, and regardless of whether it is primed or not.
After a random well is activated and the reward is collected, the home well is armed, and if the agent reaches it and collects the reward, a new random well is armed and so forth, with no maximum number of cycles.
Episodes start with the agent placed next to the home well and the priming of a random well.

\begin{figure}[htb]
\centering
  \includegraphics[width=0.95\textwidth]{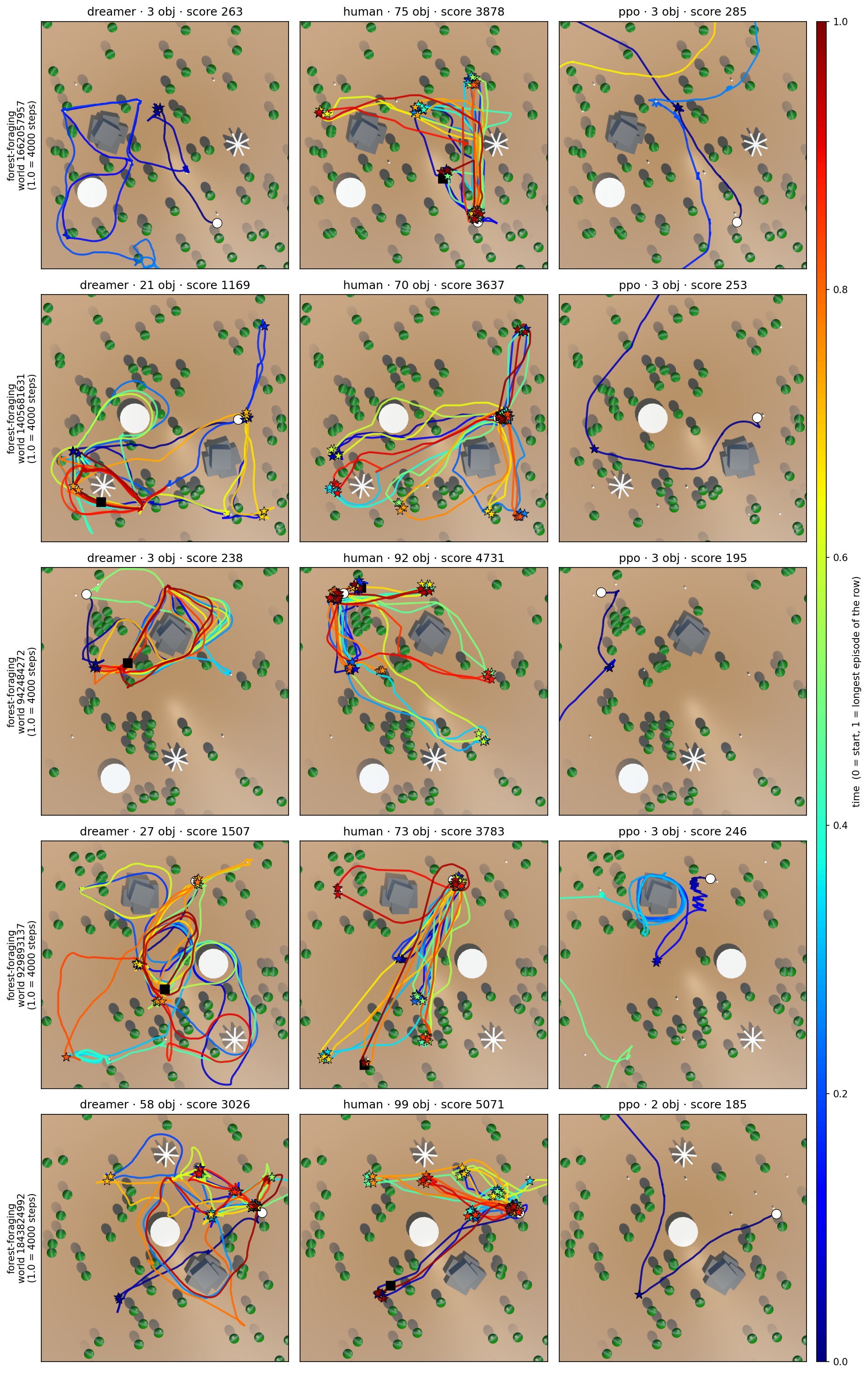}
  \caption{Visualization of eval trajectories on the \scenario{forest-foraging} unbounded well-alternation scenario. Dreamer walks outside of the main area into the infinite forest in one world, PPO in all of them.}
\label{fig:forest_foraging_trajs}
\end{figure}

\paragraph{\scenario{ratmaze-foraging}} (2000 steps, 42$\times$42\,m, $\infty$ reward objects) is a foraging (well-alternation) scenario where we replicate simplified experimental conditions seen in the rat study described in \cite{widloskiFlexibleReroutingHippocampal2022}.
In this scenario, the agent is placed within a square box with 9 well objects and randomly placed walls creating various topologies.
The generation logic makes sure that in each scenario, any 2 wells are co-reachable.
To make the scenario compatible with our 2DSDS sensing mode and reasonable RL episode lengths, we make a few changes from the original animal experiment.
Specifically, the inner walls in our scenario are not transparent, the outer walls are not textured (which the authors mainly use to give the rats a sense of direction, which we give to the agent directly through the compass observations), and the wells are not inside the ground but are standing objects, with a different semantic class than obstacles and reward objects.
Additionally, we remove the delay between well activations, and the agent can immediately move to the next well after collecting the reward from a previous one.
\paragraph{\scenario{forest-foraging}} (4000 steps, 130$\times$130\,m (containing wells and landmarks, otherwise infinite), $\infty$ reward objects) is a foraging scenario unique in that it does not contain a wall on its edges and is thus unbounded.
The environment contains uniformly scattered tree objects, and if the agent moves outside of the area with the wells, it will continue to see trees spawned around it even if it were to move infinitely far from start (and, as shown in \autoref{fig:forest_foraging_trajs}, PPO often does leave the main area).
Inside the main area, we randomly place 8 wells, and also 3 landmark objects for easier navigation, which have the same class as any other obstacle, but have unique shapes (a big cylinder, a rock formation composed form multiple boxes, and a spiky object simulating a bush or a jagged rock).
\paragraph{\scenario{4arm-submazes-foraging}} (2000 steps, 200$\times$200\,m, $\infty$ reward objects) uses the same topology and scale as \scenario{4arm-submazes}, but changes the task.
Instead of spawning 2 reward objects per room, this scenario spawns one well per every room (aside for the central one), for a total of 18 wells.

\section{Additional Metrics and RL Training Curves}
\begin{figure}[htb]
\centering
\includegraphics[width=1.0\textwidth]{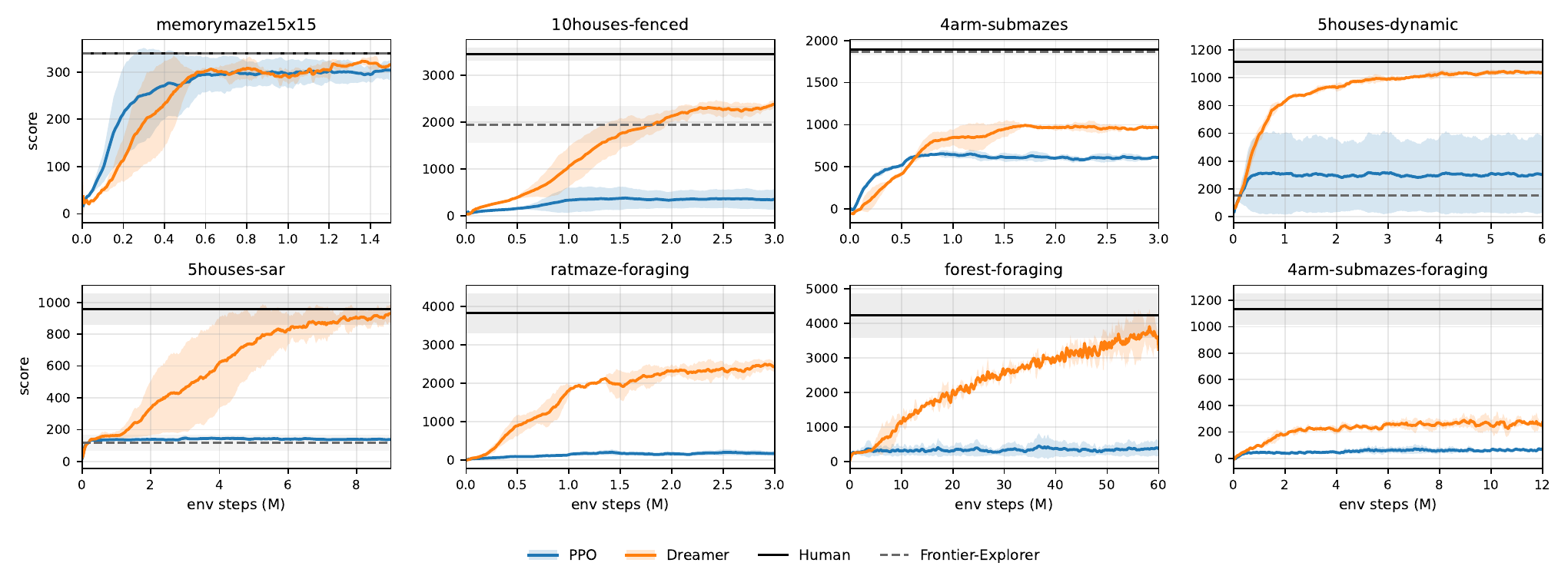}
\includegraphics[width=1.0\textwidth]{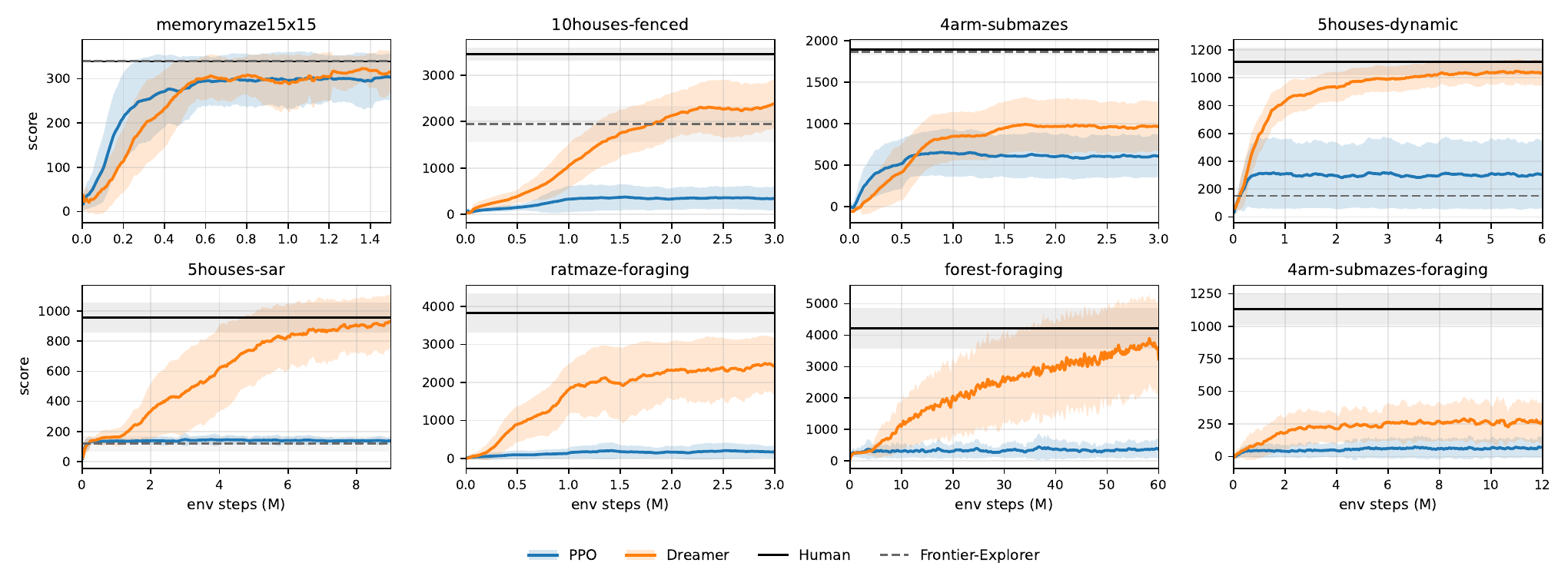}
  \caption{Full training curves of for each of the presented worlds. In all figures, the line represents mean return per episode averaged across training seeds. Top 2 rows show the standard deviation of the mean returns between individual training seeds. Bottom 2 rows show the deviation across \textit{episodes} averaged over training seeds, which shows that the baseline performance varies drastically between individual generated worlds within any scenario, and corresponds to the high eval-world score deviation in the evaluations in \autoref{sec:8worlds_experiment}}
\label{fig:training_curves_full}
\end{figure}

Training curves for PPO and Dreamer in the 8 presented scenarios are shown in \autoref{fig:training_curves_full}.

\begin{table}[!htb]
\centering
\small
\setlength{\tabcolsep}{4pt}
\renewcommand{\arraystretch}{1.15}
\providecommand{\cellEval}[1]{#1}
\providecommand{\cellTrain}[1]{{\color{blue}#1}}
\providecommand{\cellPartial}[1]{{\color{orange}#1}}
\providecommand{\cellMissing}{\textcolor{red}{--}}
\providecommand{\cellNA}{--}
\begin{tabular}{ll lll}
\toprule
\textbf{World} & \textbf{Method} & Score & OBJ & Coverage [$10^{2}$\,m$^2$] \\
\midrule
\multirow{4}{*}{memorymaze15x15} & PPO & $309_{\cellEval{\pm 39}}$ & $5.5_{\cellEval{\pm 0.7}}$ & $36.4_{\cellEval{\pm 2.7}}$ \\
 & Dreamer & $320_{\cellEval{\pm 26}}$ & $5.7_{\cellEval{\pm 0.5}}$ & $35.9_{\cellEval{\pm 2.2}}$ \\
 & Frontier-Explorer & $\mathbf{339}_{\cellEval{\pm 1}}$ & $6.0_{\cellEval{\pm 0.0}}$ & $39.2_{\cellEval{\pm 0.7}}$ \\
 & Human & $\mathbf{339}_{\cellEval{\pm 1}}$ & $6.0_{\cellEval{\pm 0.0}}$ & $38.9_{\cellEval{\pm 0.7}}$ \\
\cmidrule(lr){1-5}
\multirow{4}{*}{10houses-fenced} & PPO & $366_{\cellEval{\pm 303}}$ & $3.7_{\cellEval{\pm 3.6}}$ & $190.0_{\cellEval{\pm 159.8}}$ \\
 & Dreamer & $2292_{\cellEval{\pm 481}}$ & $37.9_{\cellEval{\pm 8.4}}$ & $401.0_{\cellEval{\pm 92.6}}$ \\
 & Frontier-Explorer & $1947_{\cellEval{\pm 389}}$ & $30.2_{\cellEval{\pm 8.5}}$ & $436.9_{\cellEval{\pm 40.2}}$ \\
 & Human & $\mathbf{3452}_{\cellEval{\pm 134}}$ & $58.8_{\cellEval{\pm 2.7}}$ & $512.6_{\cellEval{\pm 10.7}}$ \\
\cmidrule(lr){1-5}
\multirow{4}{*}{4arm-submazes} & PPO & $504_{\cellEval{\pm 210}}$ & $9.3_{\cellEval{\pm 4.3}}$ & $47.8_{\cellEval{\pm 36.4}}$ \\
 & Dreamer & $777_{\cellEval{\pm 160}}$ & $14.9_{\cellEval{\pm 3.1}}$ & $43.3_{\cellEval{\pm 8.9}}$ \\
 & Frontier-Explorer & $1868_{\cellEval{\pm 44}}$ & $35.4_{\cellEval{\pm 0.9}}$ & $98.5_{\cellEval{\pm 1.5}}$ \\
 & Human & $\mathbf{1895}_{\cellEval{\pm 3}}$ & $36.0_{\cellEval{\pm 0.0}}$ & $99.5_{\cellEval{\pm 3.5}}$ \\
\midrule
\multirow{4}{*}{5houses-dynamic} & PPO & $377_{\cellEval{\pm 282}}$ & $5.1_{\cellEval{\pm 5.8}}$ & $132.6_{\cellEval{\pm 30.9}}$ \\
 & Dreamer & $984_{\cellEval{\pm 120}}$ & $16.9_{\cellEval{\pm 2.4}}$ & $152.9_{\cellEval{\pm 5.7}}$ \\
 & Frontier-Explorer & $152_{\cellEval{\pm 0}}$ & $0.0_{\cellEval{\pm 0.0}}$ & $151.6_{\cellEval{\pm 0.2}}$ \\
 & Human & $\mathbf{1114}_{\cellEval{\pm 100}}$ & $19.2_{\cellEval{\pm 1.8}}$ & $159.4_{\cellEval{\pm 10.1}}$ \\
\cmidrule(lr){1-5}
\multirow{4}{*}{5houses-sar} & PPO & $142_{\cellEval{\pm 45}}$ & $0.2_{\cellEval{\pm 0.8}}$ & $133.2_{\cellEval{\pm 19.7}}$ \\
 & Dreamer & $864_{\cellEval{\pm 196}}$ & $14.8_{\cellEval{\pm 3.7}}$ & $135.0_{\cellEval{\pm 20.9}}$ \\
 & Frontier-Explorer & $118_{\cellEval{\pm 49}}$ & $0.0_{\cellEval{\pm 0.0}}$ & $118.0_{\cellEval{\pm 49.0}}$ \\
 & Human & $\mathbf{957}_{\cellEval{\pm 99}}$ & $16.6_{\cellEval{\pm 1.9}}$ & $132.2_{\cellEval{\pm 11.6}}$ \\
\midrule
\multirow{3}{*}{ratmaze-foraging} & PPO & $470_{\cellEval{\pm 525}}$ & $9.3_{\cellEval{\pm 10.4}}$ & $12.9_{\cellEval{\pm 6.2}}$ \\
 & Dreamer & $2883_{\cellEval{\pm 256}}$ & $57.7_{\cellEval{\pm 5.1}}$ & $17.6_{\cellEval{\pm 0.0}}$ \\
 & Human & $\mathbf{3827}_{\cellEval{\pm 515}}$ & $76.2_{\cellEval{\pm 10.3}}$ & $17.6_{\cellEval{\pm 0.0}}$ \\
\cmidrule(lr){1-5}
\multirow{3}{*}{forest-foraging} & PPO & $335_{\cellEval{\pm 269}}$ & $4.8_{\cellEval{\pm 5.0}}$ & $99.5_{\cellEval{\pm 29.9}}$ \\
 & Dreamer & $1619_{\cellEval{\pm 1370}}$ & $29.9_{\cellEval{\pm 27.2}}$ & $136.0_{\cellEval{\pm 20.6}}$ \\
 & Human & $\mathbf{4220}_{\cellEval{\pm 639}}$ & $81.8_{\cellEval{\pm 12.9}}$ & $130.6_{\cellEval{\pm 5.9}}$ \\
\cmidrule(lr){1-5}
\multirow{3}{*}{4arm-submazes-foraging} & PPO & $94_{\cellEval{\pm 79}}$ & $1.0_{\cellEval{\pm 1.5}}$ & $46.0_{\cellEval{\pm 45.6}}$ \\
 & Dreamer & $272_{\cellEval{\pm 131}}$ & $4.5_{\cellEval{\pm 2.5}}$ & $49.2_{\cellEval{\pm 19.5}}$ \\
 & Human & $\mathbf{1133}_{\cellEval{\pm 119}}$ & $21.6_{\cellEval{\pm 2.5}}$ & $55.3_{\cellEval{\pm 5.7}}$ \\
\bottomrule
\end{tabular}

  \caption{Additional metrics for the evaluation described in \autoref{sec:8worlds_experiment} and in \autoref{tab:results}. OBJ signifies number of collected reward objects, coverage is the area explored by the agent's sensors in the order of $10^2$, since this is equal to the contribution of the coverage to the overall score (since we give a reward of 0.01 for each covered square meter).}
\label{tab:results_appendix}
\end{table}

\begin{figure}[htb]
\centering
  \includegraphics[width=0.95\textwidth]{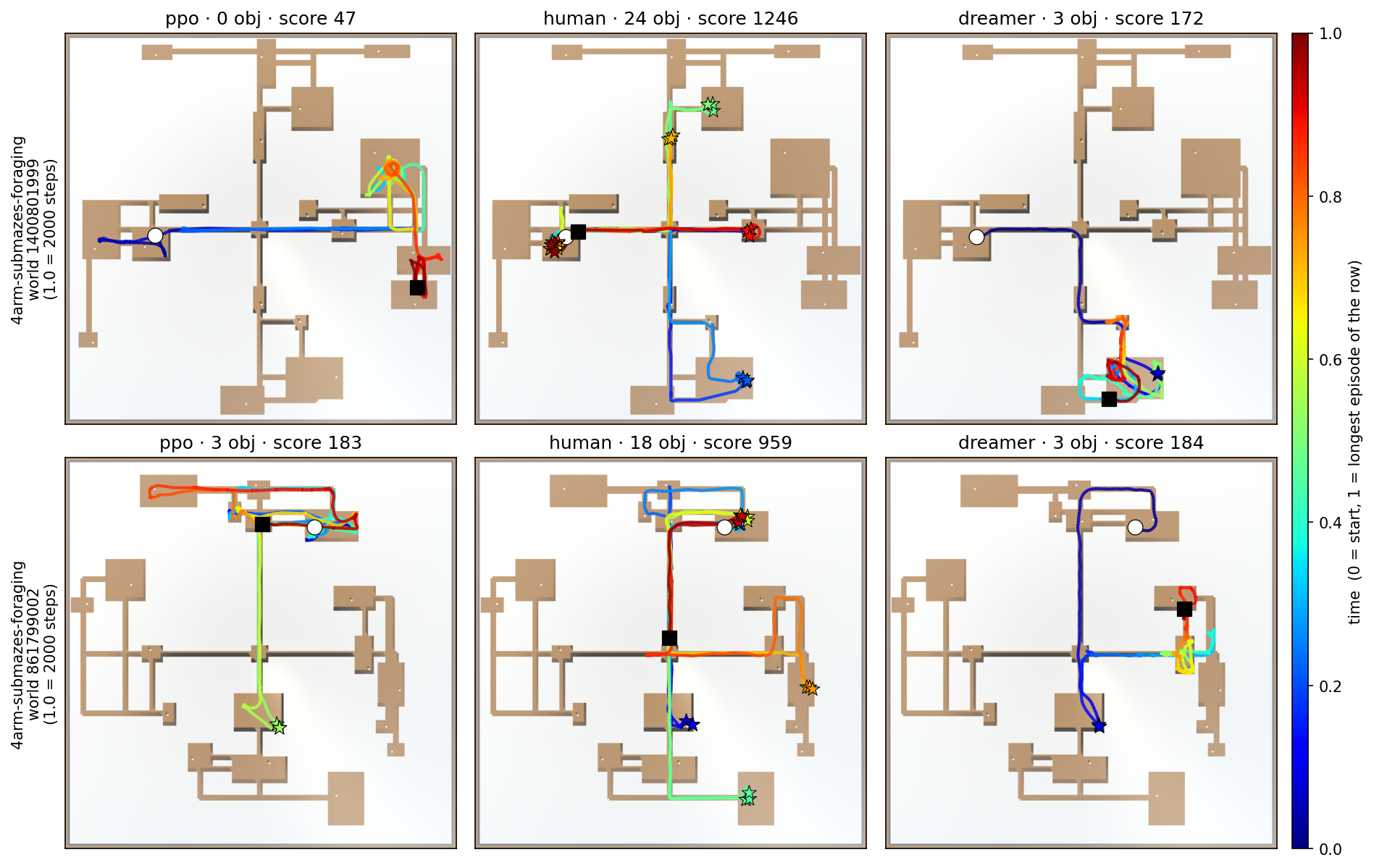}
  \caption{Visualization of eval trajectories on the \scenario{4arm-submaze-foraging} well-alternation scenario. The human player can perform 3 or more random-home alternations while both PPO and Dreamer struggle with learning to return home after finding the cued random well.}
\label{fig:ortho_foraging_trajs}
\end{figure}

\section{Extra Trajectory Visualizations}
\label{sec:trajectories}
\begin{figure}[htb]
\centering
\includegraphics[width=1.0\textwidth]{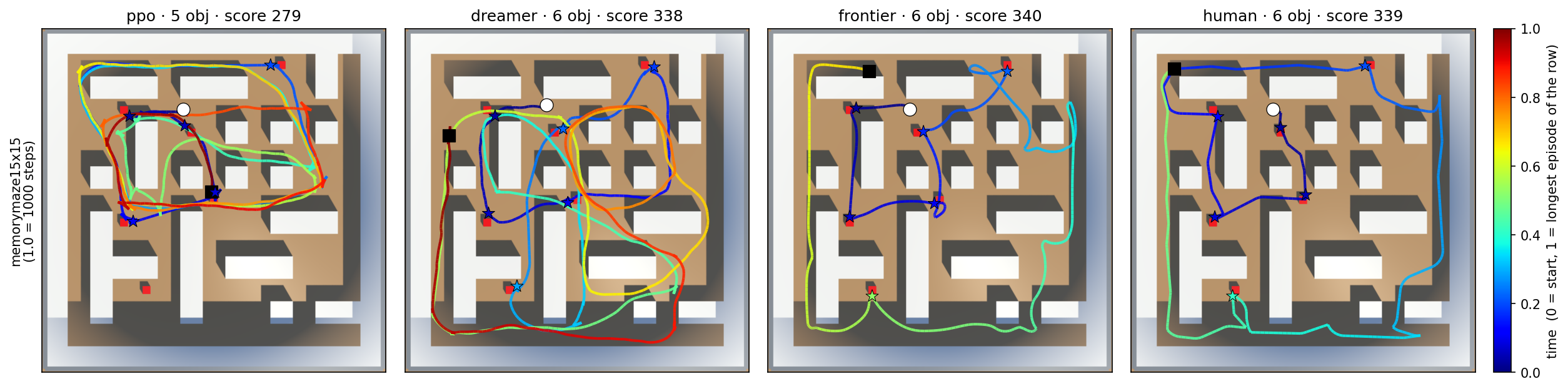}
  \caption{Trajectory visualizations for the \scenario{memorymaze15x15} world, showing that it can be easily fully covered in the multi-object search task by even a model-free, memoryless PPO agent. This suggests that with enough time (since the human and the frontier explorer were able to cover it sometimes in 50\% of the episode steps), this environment is not challenging in terms of map-based reasoning or memory (although the original \textit{ordered}-search task is, as the authors present, shown to be a good benchmark of memory).}
\label{fig:trajs_memorymaze}
\end{figure}

Visualizations of the eval trajectories from the most informative runs are shown in \autoref{fig:trajs_memorymaze}, \autoref{fig:forest_foraging_trajs} and \autoref{fig:ortho_foraging_trajs}.
These show some of the runs recorded as the eval runs in the experimental section.

\section{Frontier-Explorer Implementation Details}
\label{sec:frontier-imple}
The frontier explorer is implemented as a ROS2 node which runs in lockstep with the simulator (meaning that the simulator waits for the node's output target velocity, to be fair since RL and the Human also have theoretically unbounded time to compute the next step).
The node builds a classic occupancy grid using the same 2DSDS sensing data  as if it was a 2D LiDAR, however without marking hits of reward objects as obstacles in the map (the semantic class label for them is supplied to the node), and using the pose observations that RL policies get as well in the multi-object search scenarios.
The occupancy grid has a resolution of 0.5\,m per cell.
Frontiers~\citep{paper_frontier_grandpa} are computed as free cells adjacent to unknown cells, grouped into clusters and clusters under 5 cells are discarded.
When the node selects a goal (either a frontier or a reward object), it plans paths using A*, with an inflation radius of 0.4\,m from obstacles and plans only through free space.
These paths are then followed using a standard carrot-on-a-stick path-following approach which calculates the output target velocities.
The node alternates between 2 modes --- frontier exploration and reward-object collection.
Frontier-exploration is the default mode and it repeatedly detects frontier clusters, selects the shortest one (based on a Dijstra flood-fill from the agent's position), selects that as the next navigation goal and repeats this process.
If a reward object is detected, the agent switches to the reward-collection mode.
In this mode, the agent simply tries to navigate to the object's detected position and it returns to frontier exploration upon reaching the reward object position or after 10 steps of not seeing the object.
The node also finishes the episode sooner if it detects no new reachable frontiers (e.g. if it clears the whole environment).

\section{Simulation Environment Details}
\label{sec:sim_details}
\begin{figure}[htb]
\centering
\includegraphics[width=1.0\textwidth]{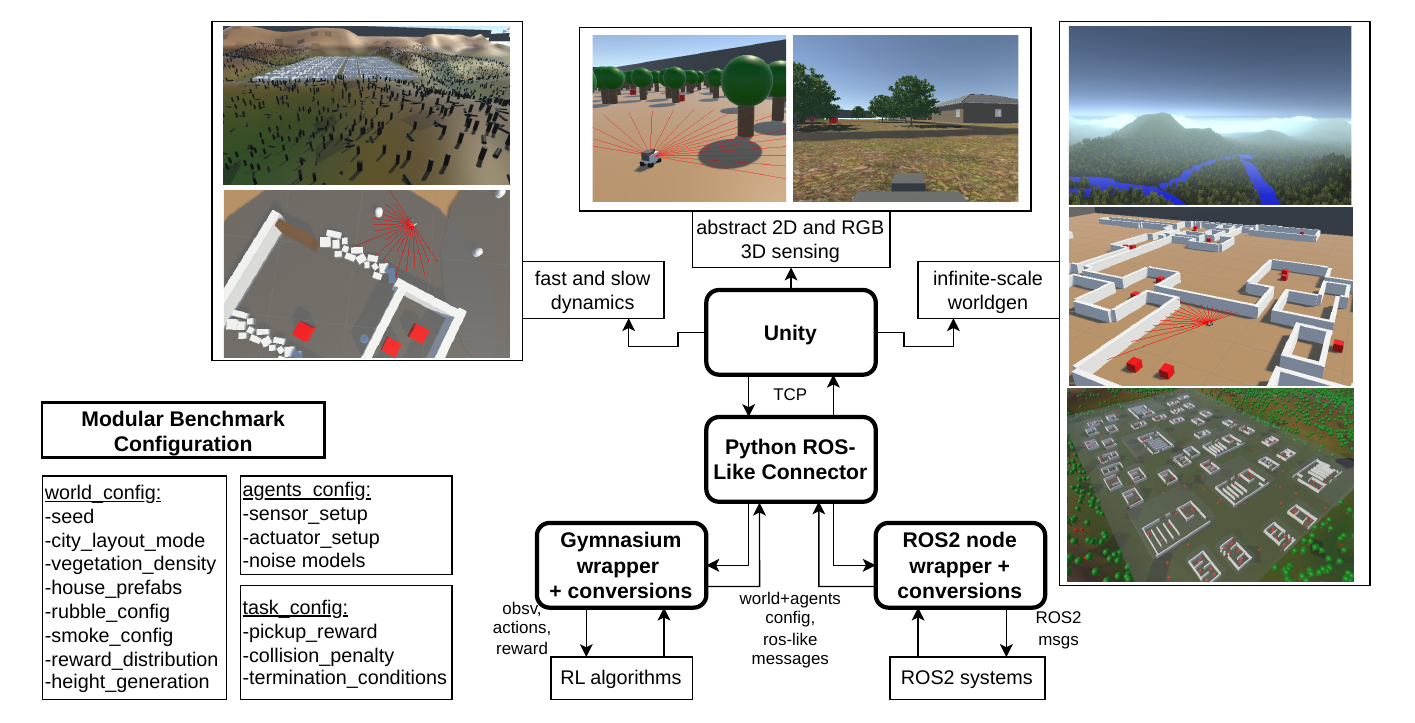}
\caption{\bench environment architecture overview. }
\label{fig:architecture}
\end{figure}

\begin{figure}[htb]
\def\subfigwidth{0.9\linewidth}
\centering
\begin{minipage}{\subfigwidth}
  \begin{tikzpicture}
    \node[anchor=south west,inner sep=0] (a) at (0,0) {
      \adjustbox{trim={{0.15\width} {0.0\height} {0.15\width} {0.0\height}}, clip, width=1.0\linewidth}{%
        \includegraphics{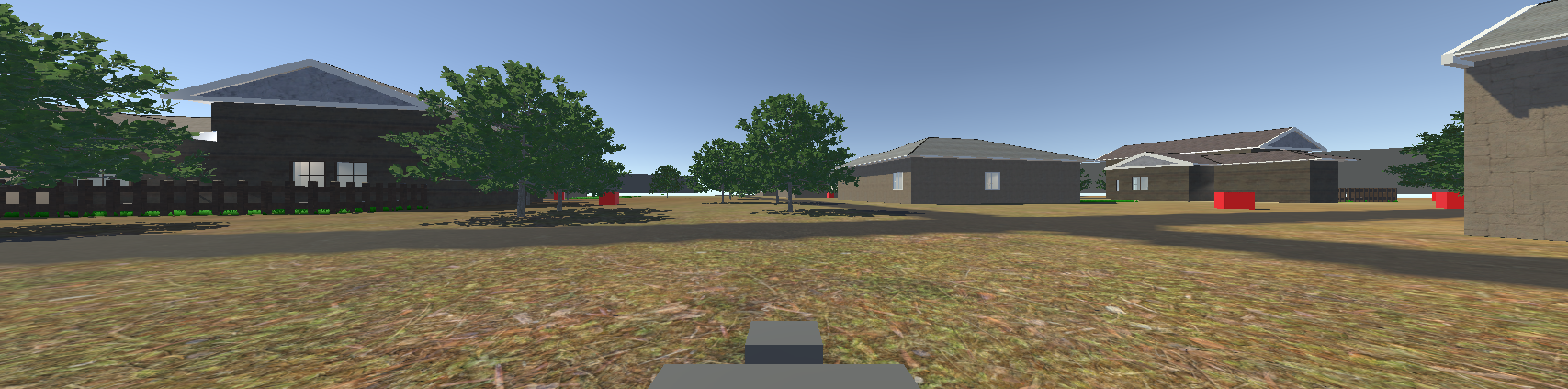}}};
  \end{tikzpicture}
\end{minipage}\hfill
\begin{minipage}{\subfigwidth}
  \begin{tikzpicture}
    \node[anchor=south west,inner sep=0] (a) at (0,0) {
      \adjustbox{trim={{0.0\width} {0.15\height} {0.0\width} {0.0\height}}, clip, width=1.0\linewidth}{%
        \includegraphics{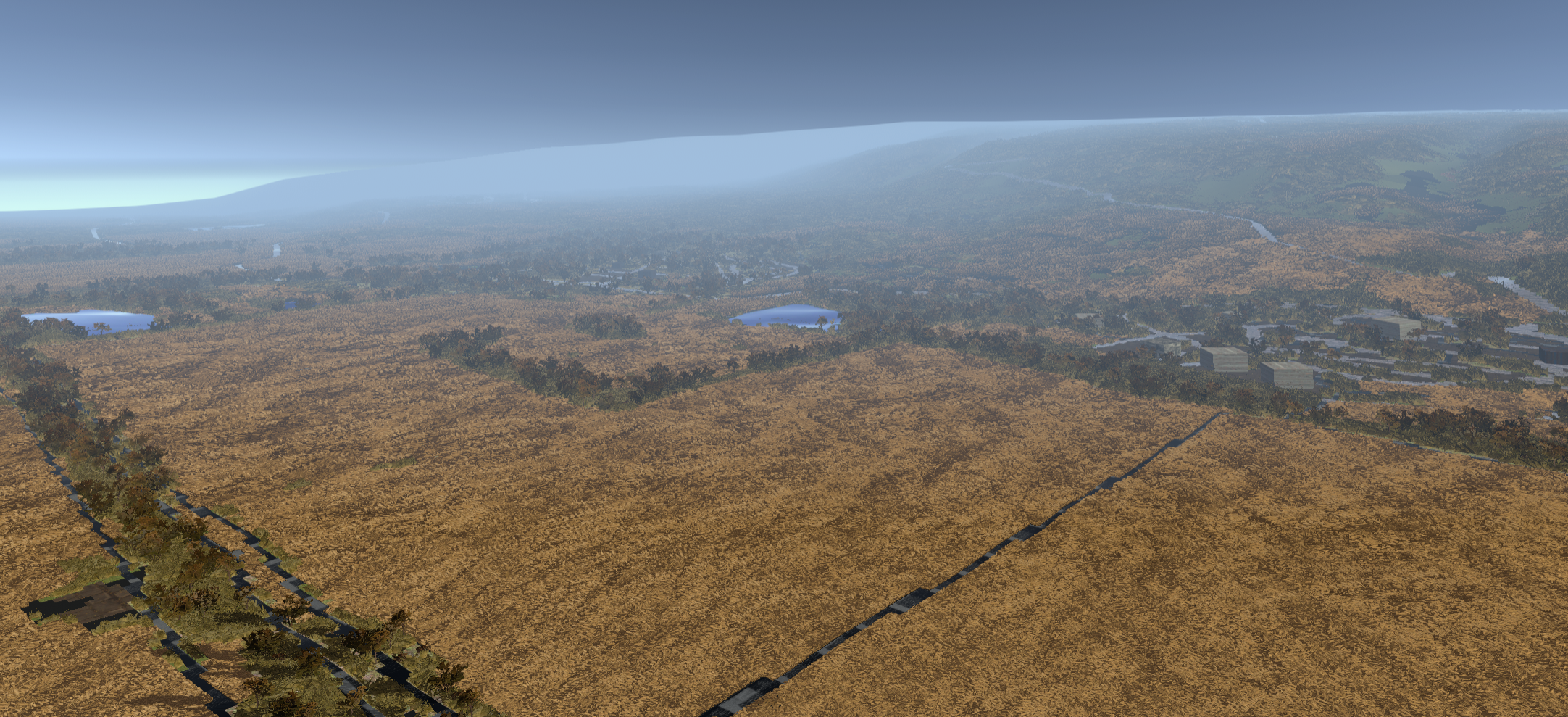}}};
  \end{tikzpicture}
\end{minipage}
  \caption{Examples of the 3D environments available in \bench in the RGB/LiDAR sensing modes, Top: a 3D variant of search-and-rescue missions in suburb-scale environments with interactible, randomized houses (with ability to swap abstract red boxes as reward objects for human models), Bottom: a recreation of the Hula valley on which the animal experiments documenting the scalability of biological cognitive maps of bats have been recorded \cite{kidnapped_bats}}
\label{fig:hero3d}
\end{figure}

Here, we describe the simulator implementation in more detail.
An overview diagram of the simulator is shown in \autoref{fig:architecture}.
We also show two of the more realistic 3D worlds which are available in \bench, but were not used in the experiments since their scale is above what current RL baselines are able to cover, and since we wanted to focus on the problems of topology, dynamics and scale while isolating away perception complexity of high-dimensional RGB observations.
These worlds are shown in \autoref{fig:hero3d} and show a recreation of the Hula valley, which is used in many publications documenting the scalability of the navigation capabilities of bats (e.g. \cite{kidnapped_bats}), and an example of a suburb environment with more realistic textures and models than the simple ones we use in the main experiments.
We designed these as a next step to help the development of future methods, once they are able to perform the search and well-alternation tasks in large and dynamic 2D scenarios in \bench, to transition to fundamentally similar but 3D worlds and eventually into real-world applications. 

\subsection{Interface and Simulation Control}
The world generation, dynamics and sensing and action simulation of \bench are implemented in C\# inside the Unity Game Engine, with a general-purpose python TCP connector interface which is then routed into both ROS2 and Gym, which map the general messages into either ROS2 messages/services or Gym observations/actions.
The way of passing data between objects in Unity and the python interface is highly inspired by the ROS architecture of messages flowing in topics, with nodes able to publish or subscribe to any topic.
This allows for high modularity (e.g. designing a new sensor in Unity requires only specifying a publisher object inside Unity and publishing the data on it, or designing a new actuator requires a subscriber in Unity mapping the message data into physics and writing the publishing of the action message in python on a specified topic), but we did not want to make Unity entagled with ROS2.
Messages are defined in C\# inside Unity and the python package contains tooling to automatically generate their corresponding python classes (again, free of ROS2).

The simulation is fully controlled and the physics and all dynamics are stepped by the python connector, nothing is simulated unless the user (a python application using the connector) steps the simulation.
On each step, the python connector sends all input messages (usually agent actions or world reset requests) into Unity through TCP, which are routed to subscribers in Unity objects.
Afterwards, one step of the simulation (including physics) is performed, and then messages from all publishers within Unity are collected and sent back to python.
For training, the simulation runs in headless mode with no screen output and the python interface allows running multiple Unity processes in parallel on different ports, which is how we implement multiple environments in our RL training.
When evaluating human performance, we run in normal mode, with the Human UI visualized in \autoref{fig:obs_modes} dislayed on screen and the process accepting WASD/Space inputs.

\subsection{Modular World Generation and World Scalability}
To enable easy extension with new generation mechanisms and to reduce the headache caused by designing world generation order, \bench's world generation architecture is designed as a set of Generator modules, each of which can provide or require some type of data.
The order of the generation determined based on these requirements upon world generation, which allows easily adding new generation mechanics (e.g. if we wanted to add generation of realistic roads, we would create a RoadGenerator module and make it require Height data and provide VegetationClearance data which is then subscribed by a VegetationGenerator).

The generation is then activated by agents --- as an agent moves through the world, it \textit{requests} world-data in chunks surrounding the agent as the agent moves through the world, and unloaded when the agent leaves the area, similar to Minecraft.
This triggers individual generators by walking the generation dependency graph.
However, since we want to be able to generate not just chunk-sized objects, we also use the concept of 'structures', which are oriented rectangular objects in the world, and the generation of their contents is triggered when loading of a chunk overlapping them is requested.
While the simulator theoretically allows generating structures infinitely, in the presented experiments we use bounded worlds (and in the case of \scenario{forest-foraging}, we purposefully do not want the landmarks or wells to be outside of the 'content area'), and thus the high-level structures are placed once on world generation.
The high-level structures (e.g. cities/mazes) can then spawn lower-level structures (houses, rooms etc.) as the agent gets near them.
Grid-based data does not require any structures to exist, but can be influenced by structures, for example vegetation data can be influenced by houses.
The user then can define worlds as a set of a high-level structures they want within a world, (e.g. 5 houses, 1 maze etc.) and then specify other parameters which guide the lower levels of the generation (e.g. specifying the number of rooms within a maze, types of prefabs allowed in a city, or allowing reward spawning only within rooms labeled 'reward-room' or spawning wells instead of reward objects).

\section{Hyperparameters for Training}
\label{app:hparams}
Both RL baselines are trained on the same Gym environment with 4 parallel simulator instances and identical observations and action space.
For parameters, PPO uses the Stable-Baselines3 v2.7.1 defaults without modification (\texttt{MultiInputPolicy}: flattened and concatenated observations, separate $2\times64$ tanh MLP actor and critic), trained on CPU only (16 threads per run).
\autoref{tab:hparams} lists the DreamerV3 hyperparameters that differ from the defaults of the official JAX implementation. 
Everything not listed is the default.
Since the observations are vectors, only the MLP branches of the encoder and decoder are used.

\begin{table}[htb]
\centering
\small
\begin{tabular}{lp{0.62\linewidth}}
\toprule
\textbf{Parameter} & \textbf{Value (default)} \\
\midrule
Model size & \texttt{size12m}: RSSM deter 2048, hidden 256, 16 classes, MLP units 256 (\texttt{size200m}) \\
Batch size & 8 (16) \\
Sequence length & 48 (64) \\
Replay context & 16 (1) \\
Replay buffer & $10^6$ steps ($5\times10^6$) \\
Parallel envs & 4 (16) \\
Compute dtype & float32 (bfloat16) \\
\midrule
\multicolumn{2}{l}{\textit{Memory variants (\autoref{sec:adaptive_experiment}), relative to the above}} \\
\midrule
bl256 & Sequence length 256 \\
consec4 & 4 consecutive 48-step chunks with carried recurrent state \\
TBTT & Sequential replay with state carried across episodes, replay context 0, following~\cite{memorymaze} \\
\bottomrule
\end{tabular}
\caption{DreamerV3 hyperparameters that differ from the official implementation's defaults (defaults in parentheses). PPO uses the Stable-Baselines3 defaults.}
\label{tab:hparams}
\end{table}


\newpage

\end{document}